%% file: main.tex
\pdfoutput=1
\documentclass[nonacm,sigplan]{acmart}

\usepackage[]{hyperref}

\usepackage{microtype}
\usepackage{graphicx}
\usepackage{subfigure}
\usepackage{booktabs} 
\usepackage{mathtools}
\usepackage{multirow}
\usepackage{float}
\usepackage{enumitem}
\usepackage{cleveref}
\usepackage{wrapfig}
\usepackage{array}
\usepackage[]{algorithm2e}

\usepackage[frozencache,cachedir=.]{minted}
\definecolor{lgray}{gray}{0.95}

\newcommand{\system}{\text{SENSEi}}

\newcommand{\mycomment}[1]{}

\begin{document}

\title{SENSEi: Input-Sensitive Compilation for Accelerating GNNs}

\author{Damitha Lenadora}
\affiliation{%
  \institution{University of Illinois at Urbana-Champaign}
  \country{United States}}
\email{damitha2@illinois.edu}

\author{Vimarsh Sathia}
\affiliation{%
  \institution{University of Illinois at Urbana-Champaign}
  \country{United States}}
\email{vsathia2@illinois.edu}

\author{Gerasimos Gerogiannis}
\affiliation{%
  \institution{University of Illinois at Urbana-Champaign}
  \country{United States}}
\email{gg24@illinois.edu}

\author{Serif Yesil}
\affiliation{%
  \institution{University of Illinois at Urbana-Champaign}
  \country{United States}}
\authornote{Now at NVIDIA. He can be reached at syesil@nvidia.com.}
\email{syesil2@illinois.edu}

\author{Josep Torrellas}
\affiliation{%
  \institution{University of Illinois at Urbana-Champaign}
  \country{United States}}
\email{torrella@illinois.edu}

\author{Charith Mendis}
\affiliation{%
  \institution{University of Illinois at Urbana-Champaign}
  \country{United States}}
\email{charithm@illinois.edu}

\begin{abstract}
Over the years, many frameworks and optimization techniques have been proposed to accelerate graph neural networks (GNNs). 
Compared to the optimizations explored in these systems, we observe that different matrix re-associations of GNN computations lead to novel input-sensitive performance behavior.
We leverage this observation to propose \system{}, a system that \textit{exposes} different sparse and dense matrix primitive compositions based on different matrix re-associations of GNN computations and \textit{selects} the best among them based on input attributes. \system{} executes in two stages: (1) an offline compilation stage that enumerates all valid re-associations leading to different sparse-dense matrix compositions and uses input-oblivious pruning techniques to prune away clearly unprofitable candidates and (2) an online runtime system that explores the remaining candidates and uses light-weight cost models to select the best re-association based on the input graph and the embedding sizes on a given hardware platform.
On a wide range of configurations, \system{} achieves speedups of up to $2.012\times$ and $1.85\times$ on graph convolutional networks and up to $6.294\times$ and $16.274\times$ on graph attention networks, on GPUs and CPUs respectively. 
We also show that its technique generalizes to GNN variants, including those that require sampling.
Furthermore, we show that \system{}'s techniques are agnostic to the underlying GNN system, and can be used to yield synergistic improvements across a diverse set of implementations. 
\end{abstract}

\maketitle 
\pagestyle{plain} 


\input{paper/introduction}

\input{paper/background_motivation}

\input{paper/system}

\input{paper/implementation}

\input{paper/primitive_compositions}

\input{paper/evaluation}

\input{paper/related_work}

\input{paper/conclusion}

\input{paper/acknowledgements}


\bibliographystyle{plain}
\bibliography{references}

\end{document}

%% file: paper/introduction.tex
\section{Introduction}

Graph Neural Networks (GNN) have gained adoption across a wide range of application domains including social media marketing~\cite{nips2018:zhang:gnn-link-predcition}, financial fraud detection~\cite{baba-fraud1,baba-fraud2}, drug discovery~\cite{iclr2019:jin:drug-discovery, acs2019:torng:gcn-drug-interaction}, and systems optimization \cite{tpu-costmodel}. However, training GNNs is expensive and usually spans multiple hours, if not days. 
As a result, there have been many efforts, including software frameworks such as DGL~\cite{iclr2019:wang:dgl}, PyG~\cite{iclr2019:fey:pyg}, SeaStar \cite{yidi2021:seastar}, and NeuGraph~\cite{atc2019:ma:neugraph}, compilers such as Graphiler~\cite{graphiler} and many systems optimization techniques~\cite{mlsys2020:jia:roc,dorylus, zhang2021understanding, bridging-gap} that aim at accelerating GNN computations. 

\begin{figure}[th]
    \centering
    \includegraphics[width=1\linewidth]{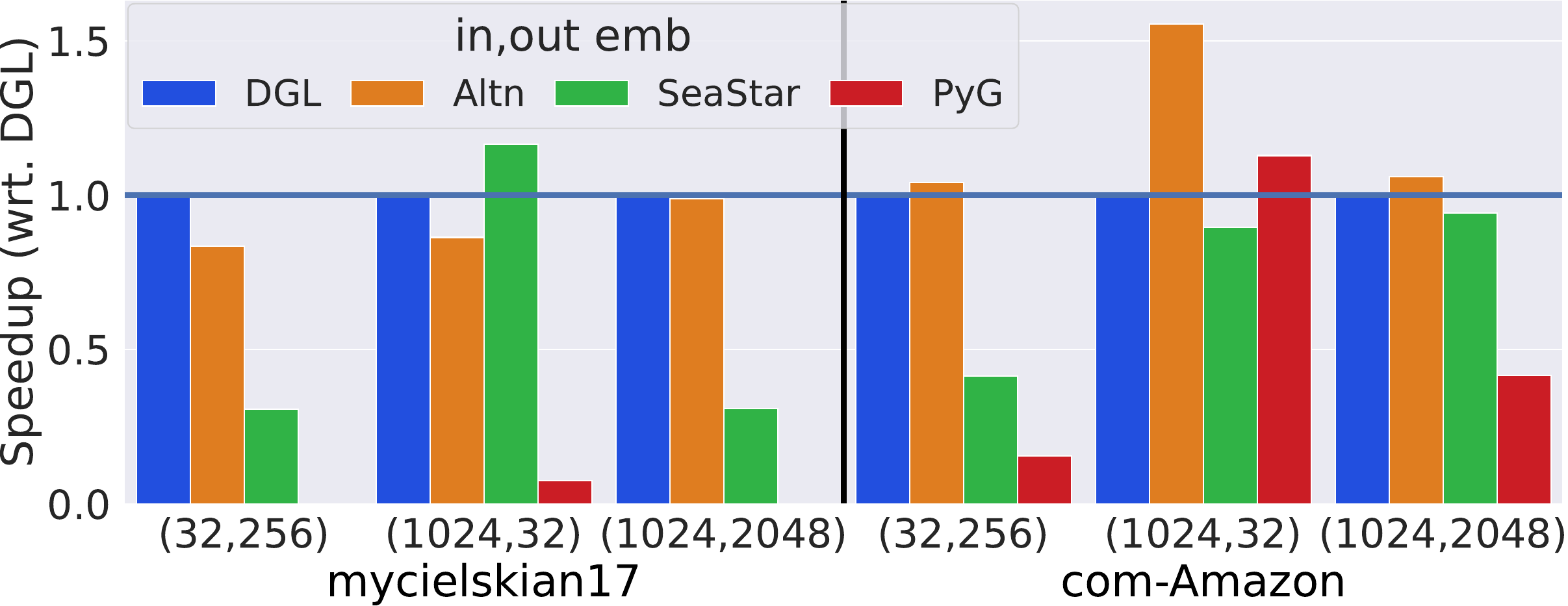}
    \caption{Comparing inference performance for multiple GCN implementations on the GPU. Performance is inconsistent across graphs and embedding sizes.  
    Empty spaces are from OOM Errors.
    (\textit{Altn} - Alternative implementation of GCN in DGL)
    } 
    \label{fig:motivation_diff}
\end{figure}

Existing systems typically model GNN computations as individual phases with input-oblivious compositions of sparse or dense matrix operations. 
In a typical GNN layer, each node collects and aggregates its neighbor node states (embeddings). 
Next, each node updates its own state using these aggregated embeddings.
The first computation is modeled as a set of sparse matrix operations, and the latter is a dense matrix multiplication. 
Apart from these phases, different GNNs perform other required computations such as normalization and edge weight (attention score) calculations that are also modeled as either sparse or dense matrix operations.
The choice of the optimal operations, as well as the optimizations applied to them, rely on inputs and configurations of the GNN model. 
However, the common practice is to build hard-coded GNN models that ignore such choices and remain the same across different inputs and configurations.
\Cref{fig:motivation_diff} shows that this results in different performance characteristics across different graphs and embedding sizes for a given system (e.g. \textit{SeaStar}). 
This phenomenon is not limited to one system but can be observed across all four GNN implementations.

\begin{figure}[th]
    \centering
    \includegraphics[width=1\linewidth]{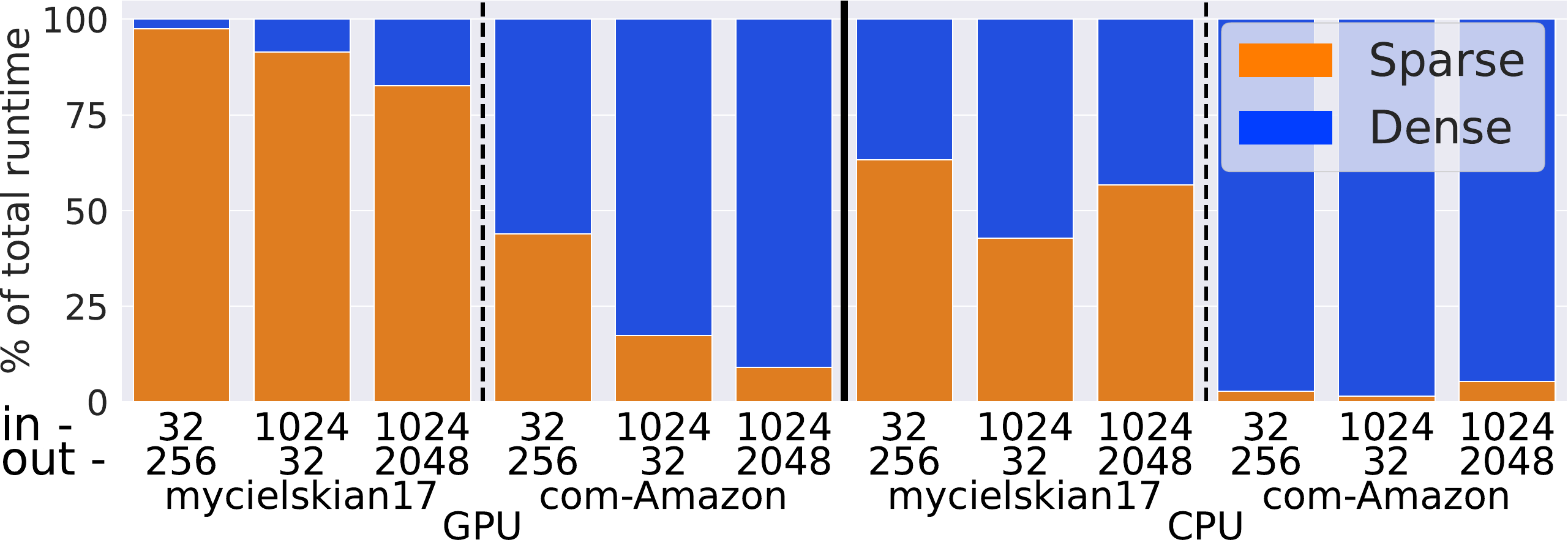}
    \caption{Percentage of computation (spare/dense) across graphs for GCN, (in, out) embedding sizes, and underlying hardware.} 
    \label{fig:motivation_bottle}
\end{figure}

Depending on the input graph, embedding sizes, and underlying hardware, the bottleneck of the GNN computation can shift between the sparse matrix and the dense matrix operations. 
This is clearly seen in Figure \ref{fig:motivation_bottle} where for \textit{mycielskian17} sparse computations are the bottleneck, while dense is the bottleneck for \textit{com-Amazon}. 
Since hard-coded models use the same sparse-dense matrix primitive compositions across all inputs, they cannot adapt to the shifts in the bottleneck. 
This is one reason why we see different performance characteristics across inputs.
Changing the primitive compositions in a system, with respect to the input, can potentially improve runtime performance.
This is seen \Cref{fig:motivation_diff} where \textit{Altn} shows significant speedups for the $(1024,32)$ embedding size in \textit{com-Amazon}.
\textit{Altn} is an implementation of GCN which uses a different primitive composition than DGL.

We use this observation to build \system{}, a system that explores semantically equivalent matrix re-associations that lead to different sparse-dense matrix compositions. 
It then selects the best, given the input graph and the embedding sizes.
For example, the computation $A{\cdot}B{\cdot}C$, can either be associated as $(A{\cdot}B){\cdot}C$, $A{\cdot}(B{\cdot}C)$ or $(A{\cdot}B{\cdot}C)$. 
In each of these associations, the primitives used, as well as their order, would differ based on the characteristics of the matrices (like $dense$ or $sparse$ and so on). 
\system{} executes in two stages: (1) an offline compilation stage that enumerates all valid matrix re-associations and prunes clearly unprofitable candidates using input-oblivious rules and (2) an online runtime system that uses lightweight cost models to select the best re-association given the input graph and the embedding size. 

Existing works perform dynamic optimizations on GNNs to achieve good performance.
GNNAdvisor ~\cite{ding-gnnadvisor} inspects the input graph and maps the computations to auto-tuned GPU kernels. 
Other works use the embedding size \cite{yan:feat_op_GNN_selection} to reorder primitives, or inspect the input graph \cite{qiu:xgboost_format_selection_GNN} to select the sparse representation to be used.  
As per our observations, three factors affect the efficiency of GNN computations -- the sparsity patterns of the input graph, the embedding sizes of each layer, and the underlying hardware. There is a joint effect of these factors on execution and computation complexity. While existing systems identify some of these factors, they do not exploit the synergy between them to drive optimizations. 
To the best of our knowledge, \system{} is the only system that leverages all these factors to automatically select the best primitive composition. It does this by using matrix-based sparse-dense primitive re-association.

Our evaluations show that \system{} achieves significant speedups of up to 26.85$\times$ on two popular GNN models: GCN and Graph Attention Network (GAT)\cite{iclr2018:velic:gat}. 
Speedups are observed across a broad range of graphs and embedding sizes on both CPUs and GPUs. 
\system{} can also generalize to other GNN variants, including models which require sampling (GraphSAGE\cite{nips2017:hamilton:graphsage}).
We conduct our main evaluation on the popular GNN framework DGL~\cite{iclr2019:wang:dgl} to demonstrate the benefit of \system{}. 
We also implement \system{} in SeaStar \cite{yidi2021:seastar} to show that the technique it introduces is not limited to a single framework. 
We make the following contributions.

\begin{itemize}[leftmargin=15pt]
    \item We present \system{}, a system, where given a GNN, is capable of automatically identifying different sparse-dense primitive compositions and selects the best based on the input using rule and data-driven techniques.
    \item We perform a case study on two widely used GNN models and introduce input-sensitive primitive compositions involving different sparse and dense matrix primitives. 
    \item We show that \system{}'s decisions lead to speedups of up to $2.012\times$ and $1.85\times$ for GCN, and $6.294\times$ and $16.274\times$ for GAT for GPU and CPU respectively.
    \item We conduct further studies to show that \system{} gives speedups for other GNN variants, is impervious to graph sampling and generalizes to other GNN frameworks.  
\end{itemize}

%% file: paper/background_motivation.tex
\section{Background}

GNN computations can be broken down into dense and sparse matrix primitives ~\cite{iclr2019:wang:dgl}.
We briefly introduce such primitives and explain when and where they are used to compute GNN models.

\subsection{Dense Matrix Primitives}
Multiple dense matrix primitives are used in GNNs due to their inherent relation to neural networks.
These are primitives where all inputs and outputs are dense matrices or vectors.
Among these primitives are element-wise computations such as non-linearity functions and matrix multiplication variants.
The latter, which includes general matrix multiplication (GEMM), is commonly found in operations such as updating features based on learned weights. 

\subsection{Sparse Matrix Primitives}

We refer to matrix operations where at least one input is a sparse matrix representation, as sparse matrix primitives. 
For GNNs, two sparse matrix primitives are commonly used. 
These are the generalized forms of sparse matrix dense matrix multiplication (SpMM) and sampled dense-dense matrix multiplication (SDDMM). 
The standard SpMM and SDDMM use $+$ and $\times$ as the addition and multiplication operators, whereas in the generalized form, the operations can come from any semi-ring \cite{davis:graphblas}.
We use $\oplus$ and $\otimes$ to signify these generalized addition and multiplication operators, and the notation $x_{i,j}$ to refer to the value at row $i$ and column $j$ of a given matrix $X$. 

\textbf{\textit{Generalized sparse matrix dense matrix multiplication (g-SpMM).}}
g-SpMM takes one input sparse matrix $A$ and multiplies that by a dense matrix $B$ to produce a dense output matrix $C$. Equation~\ref{eqn:spmm} shows how $c_{i,k}$ of $C$ is computed.
\begin{equation} \label{eqn:spmm}
c_{i,k} = \sum_{\oplus, j} a_{i,j} \otimes b_{j,k}
\end{equation}
The computation mirrors that of GEMM. However, note that when $a_{i,j}$ is zero, there is no contribution to the output. Since $A$ is sparse, the majority of its entries are zeroes. Thus, SpMM leverages the sparsity of input matrix $A$ to skip any unnecessary multiplications compared to GEMM. 

\textbf{\textit{Generalized sampled dense dense matrix multiplication (g-SDDMM).}}
g-SDDMM performs multiplication between two dense input matrices ($B$, $C$) masked by a sparse matrix $A$. This results in an output sparse matrix $D$ with a sparsity pattern matching $A$. Equation~\ref{eqn:sddmm_row} the computation of each element $d_{i,j}$ in $D$.
\begin{equation} \label{eqn:sddmm_row}
d_{i,j} = 
\begin{dcases}
    a_{i,j} \otimes \sum_{\oplus, k} b_{i,k} \otimes c_{j,k},      & \text{if } a_{i,j} \text{ not } 0\\
    0,              & \text{otherwise}
\end{dcases}
\end{equation}
In g-SDDMM, a multiply ($\otimes$) operation is performed between $b_{i,k}$ and $c_{j,k}$, for all $k$ if and only if the value of $a_{i,j}$ is non-zero. The partial products are then summed up ($\oplus$) and multiplied with $a_{i,j}$ to form the $d_{i,j}$ scalar value. 
This is again similar to GEMM with two dense input matrices. 
However, in g-SDDMM, the input sparse matrix is used as a mask to selectively compute values instead of performing the entire dense matrix multiplication.

\subsection{Matrix Primitives in GNNs}

Graph neural networks consist of two main stages: aggregation and update. 
The embeddings possessed by each node or edge as a hidden state vector are passed among neighbors to form an aggregated message during the aggregation stage. 
The messages are then transformed into updated embeddings as the GNN layer's output during the update stage.
Such computations are usually modeled as a collection of dense and sparse primitive operations~\cite{iclr2019:wang:dgl}.
Node-based aggregations are modeled as generalized sparse matrix dense matrix multiplications (g-SpMM), edge-based aggregations are modeled as generalized sampled dense-dense matrix multiplications (g-SDDMM), and update is modeled as a general matrix multiplication (GEMM). 

Other computations apart from the aforementioned, such as normalization, open up more opportunities to consider different matrix re-associations. This leads to the ability to use different sparse-dense primitive compositions when computing GNN models. 

%% file: paper/system.tex
\section{\system{} System}\label{sec:system}

We present \system{}: a \emph{GNN acceleration system} that uses matrix re-associations to exploit the input sensitivity of sparse-dense matrix primitive compositions.

\subsection{Overview}

\system{} requires minimal user interaction to setup and accelerate GNN code.
\system{} is ready to be used by simply running an initialization script.
Once this process is done, a user only needs to provide the GNN model to be accelerated as well as the inputs (graph, the input and output embedding sizes), as shown in Figure \ref{fig:sensei_code}.
\system{} then operates on the GNN model code and replaces the existing GNN model with an accelerated version. The user can then run the original code as initially intended without additional work. 
 
\begin{figure}[H]
  \begin{minted}[fontsize=\small,bgcolor=lgray,escapeinside=||]{python}
import SENSEi
graph, input_emb, output_emb = ....
model = GraphConv(..)
SENSEi(model, graph, input_emb, output_emb)
  \end{minted}
    \caption{Using \system{}.}
    \label{fig:sensei_code}
\end{figure}

\begin{figure}[htb]
    \centering
    \includegraphics[width=1\linewidth]{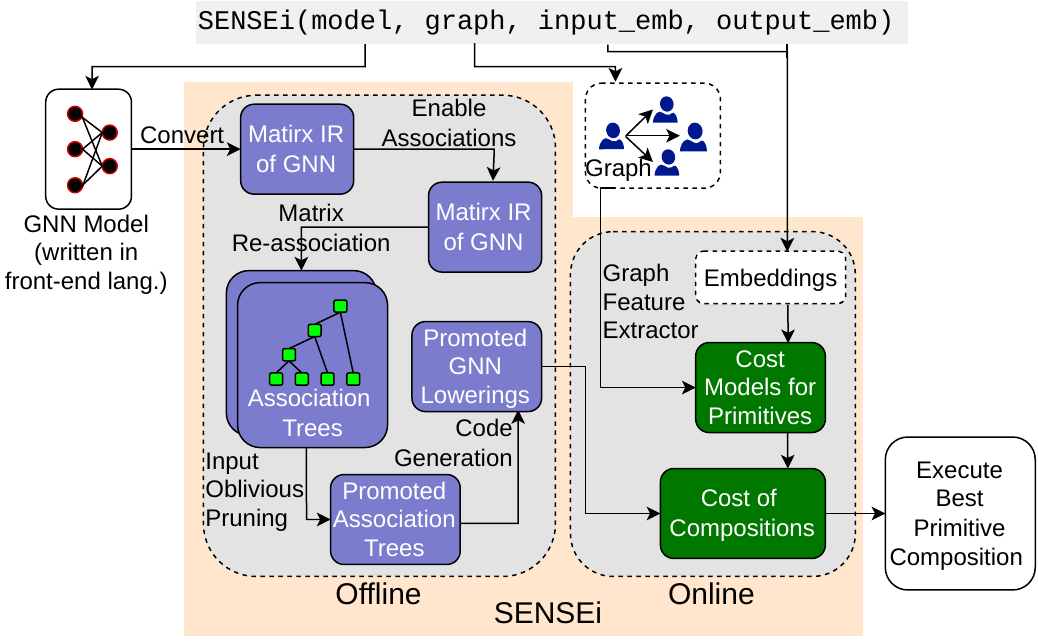} 
    \caption{Overview of the \system{} system.}
    \label{fig:overview}
\end{figure}

Figure \ref{fig:overview} presents the overall process of \system{}.
It has two main stages: (1) an offline compilation stage for generating potential candidates with different sparse-dense matrix primitive compositions and (2) an online stage for selecting the best primitive composition based on the input graph and embedding sizes.

In the offline compilation stage, \system{} first converts a GNN model written using the message passing paradigm ~\cite{iclr2019:wang:dgl} to a matrix intermediate representation (IR). 
It uses this matrix IR form to generate all possible sparse-dense matrix primitive compositions that can be used to implement the given GNN model as potential candidates using operator re-association. 
Finally, \system{} prunes away clearly unprofitable candidates using rules that are oblivious to the input and promotes the final set of candidates to the online stage to select the optimal primitive composition. 

In the online stage, \system{} selects the best primitive composition out of the promoted candidates specialized to an input graph.
It evaluates each candidate's cost using a set of \emph{input-sensitive cost models.} 
These consist of a cost model for each sparse and dense matrix primitive, which \system{} uses together to predict the final GNN execution cost. 
Compared to the pruning rules used in the offline stage, these cost models consume graph features as input in addition to the embedding sizes allowing \system{} to make more informed predictions about the relative cost of different compositions, out of which it selects the best.  

Note that this decoupled design requires the user to run only the online stage of \system{} for different input graphs. 
The offline compilation stage needs to be run only once to generate the promoted candidates.

\subsection{Matrix Representation Generation} 
\begin{figure}[htb]
    \centering
    \includegraphics[width=1\linewidth]{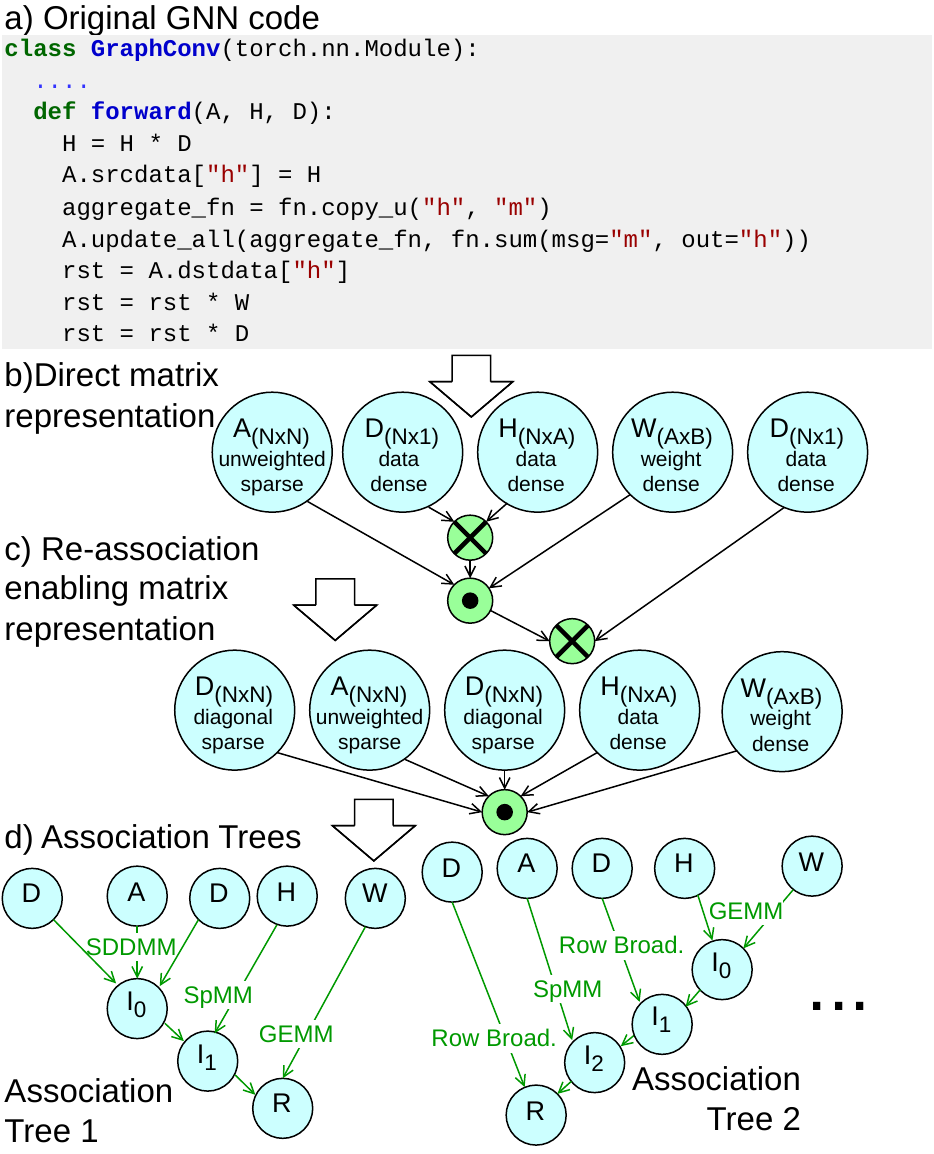}
    \caption{Association tree generation from GNN code. ($\otimes$ denotes row broadcast, $\odot$ denotes matrix multiplication)} 
    \label{fig:lowering}
\end{figure}

We use Figure~\ref{fig:lowering} as a running example to illustrate the offline stage of \system{} for a canonical GNN implementation.

\textbf{\textit{Matrix Representation.}}
We use a matrix-based intermediate representation (IR) throughout the initial stages of the offline stage of \system{} that makes it easier to generate different primitive compositions via operator re-association.
This representation is tree-based, where leaf nodes represent matrices and intermediate nodes represent matrix operations such as \textit{multiplication}, \textit{addition}, and \textit{row broadcast}. 
A leaf matrix node, $A_{(X \times Y)}^{\text{attr subattr}}$ contains information regarding the matrix size (X $\times$ Y), and the attributes (attr) and sub-attributes (subattr) described in \Cref{table:matrix_attrib}. 
Attributes provide essential information to generate primitive compositions in the latter part of \system{}'s offline compilation stage.
We note that most operators used in GNNs are associative.
As long as computations between adjacent matrices are maintained, it is possible to re-associate computations between operands which we exploit in generating different sparse-dense matrix primitive compositions for a given GNN model. 

\begin{table}[th]
\caption{Matrix attributes}
\label{table:matrix_attrib}
\begin{center}
\begin{small}
\begin{tabular}{l|l}
\toprule
Attribute & Sub-attribute  \\
\midrule
dense & data - Contains data \\
 & weight - Contains learnable weights \\
\midrule
sparse & weighted - Uses edge values  \\
& unweighted - Only uses NNZ positions \\
& diagonal - A diagonal matrix \\ 
\bottomrule
\end{tabular}
\end{small}
\end{center}
\end{table}
\textit{\textbf{{IR Generation.}}} 
\system{} first translates code written using a GNN framework's API that follows the message passing paradigm \cite{iclr2019:wang:dgl} (\Cref{fig:lowering}(a) shows an example GNN implementation), into the matrix-based IR.
We use a rule-based parser for this task (e.g., \verb|update_all| \textit{graph} operation is mapped to \textit{multiplication}).
This parser also collects type information from the original GNN code to fill in the attribute details for the leaf matrix nodes (e.g., the adjacency matrix $A$ has attribute sparse). 
For the initial code in Figure~\ref{fig:lowering}(a) this results in the matrix IR in Figure~\ref{fig:lowering}(b). 
The flattened version of the matrix IR in Figure~\ref{fig:lowering}(b) is $(A \odot ( H \otimes D) \odot W) \otimes D$ ($\otimes$ refers to a row-broadcast operation, $\odot$ refers to matrix multiplication).
Note that the IR itself has an original matrix association, and in Section~\ref{sec:comp_gen}, we will describe how we use different re-associations to generate implementations with different sparse-dense matrix compositions.

\textit{\textbf{{Post processing.}}} 
We run an additional pass on the matrix IR to uncover more opportunities for arbitrary re-association. 
For example, \emph{row-broadcast} operations can act as a barrier for re-associations. 
To eliminate this, we transform \textit{row-broadcast} operations into equivalent \textit{multiplication} operations whenever possible.
We accomplish this using a rule-based transformation that utilizes previously identified matrix attributes. 
If a broadcasted matrix in the \textit{row-broadcast} operation has the same number of rows as the only sparse matrix in the original code, then that matrix can be transformed into a \textit{diagonal sparse matrix} with the computation being changed to \textit{multiplication}.
By applying this transformation, \system{} converts \Cref{fig:lowering}(b) to have only multiplications. 
This collapses the matrix IR into a single multi-operand multiplication node as shown in \Cref{fig:lowering}(c).
This now allows for arbitrary re-associations between all matrix nodes.

\subsection{Generating Sparse-Dense Matrix Compositions}\label{sec:comp_gen}

\textit{\textbf{{Association Tree Representation.}}} 
\system{} converts the matrix IR into \emph{association trees} to concretely represent sparse or dense matrix primitives in place of matrix operations. A given association tree represents one possible matrix re-association. \emph{Association trees} reuse the same matrix nodes used in the matrix IR as leaves. Internal nodes of the tree represent intermediate results, and edges represent sparse or dense matrix primitives. For example, consider the association tree $1$ shown in Figure~\ref{fig:lowering}(d).

\textit{\textbf{{Association Tree Generation.}}}
\system{} generates \textit{association trees} for all valid matrix re-associations for a computation represented in matrix IR. 
Each association tree uses its own set of sparse and dense matrix primitives, and different associations lead to different primitive compositions. 
Figure~\ref{fig:lowering}(d) shows two possible associations for the matrix IR.
In this example, association tree $1$ represents the flattened computation $((D \odot  A \odot D) \odot H) \odot W$. 
The association of $(D \odot  A \odot D)$ yields a \textit{SDDMM} primitive.  
However, association tree $2$ does not take this route, resulting in a primitive composition that uses row-broadcasts instead. 

Algorithm~\ref{alg:tree_gen} shows how \system{} recursively enumerates all possible association trees in a depth-first manner. 
It accepts the current matrix IR and an association tree as input. Then, \system{} traverses the current matrix IR to find association candidates that involve only leaf nodes (line 2). 
\system{} uses a set of rules to find such candidates based on the connected operation, attributes, and sub-attributes of the matrix IR nodes and also to specify which dense or sparse matrix primitive to use for that association. We present a few (for brevity) of these rules as follows. 
\begin{itemize}[noitemsep]
    \item A multiplication between $3$ matrices where $2$ are \emph{sparse diagonal} and the other is \emph{sparse weighted/unweighted}, can be mapped to a SDDMM primitive. 
    To achieve this, the diagonal sparse matrices are transformed into dense vectors and then used as the dense inputs to the \textit{SDDMM} primitive ($D,A,D$ nodes in Association Tree 1 in Figure \ref{fig:lowering}).
    \item A multiplication between a matrix with a \emph{sparse unweighted} attribute, and another matrix with a \emph{dense} attribute, can be mapped to a cheaper \textit{SpMM} with a reduced memory footprint that does not use the values of the matrix.
\end{itemize}
For every candidate in this list, \system{} creates a new tree (\emph{newTr}) by augmenting the current tree with an internal node corresponding to the intermediate result of the association and edges annotated with the sparse or dense matrix primitive that connect this node with the associated nodes. Simultaneously, it creates a new matrix IR (\emph{newIR}) that replaces the set of associated nodes with this new internal node (line 7) in the current matrix IR. For example, consider the association tree 1 in Figure~\ref{fig:lowering}(d). During its creation, nodes $D, A, D$ are first associated with the operation SDDMM (edge) to produce the intermediate result $I_0$. Simultaneously, \system{} also creates a new matrix IR that replaces nodes $D, A, D$ with $I_0$. This tree is further expanded recursively until no more association candidates can be found. Since we enumerate all candidate associations (line 6), this algorithm produces a forest of all valid association trees as output. Once the trees are fully generated, \system{} scans all trees to exploit any opportunities to reuse computed values (common sub-expression elimination in the compiler domain). 
\RestyleAlgo{ruled}
\begin{algorithm}[th]
\SetNoFillComment
\LinesNumbered
\caption{Generation of association trees}\label{alg:tree_gen}
\KwIn{Matrix IR of GNN code ($mIR$), current association tree ($tr$)}
\KwOut{Association Forest ($fr$)}
\SetKwFunction{FMain}{generateTree}
    \SetKwProg{Fn}{Function}{:}{}
    \Fn{\FMain{$mIR,tr$}}{
        \tcc{get resolved associations}
        
        $ candidates  \longleftarrow getCandidates(mIR) $;    
        
        \eIf{no candidates}
            {
                $fr.add(tr)$ \tcc{add tree to forest}
            }
            {
            \ForEach{$ cand \in candidates $}
                {
                    $newIR, newTr \longleftarrow apply(mIR, tr, cand)$
                    
                    $generateTree(newIR, newTr)$
                }
            }
}

\end{algorithm}

\textbf{\textit{Pruning Associations.}}
\system{} prunes unprofitable candidates irrespective of the input from the generated forest of association trees.
We find sets of unprofitable candidates under two scenarios: 1) the input embedding size is larger than the output ($<$), and 2) vice-versa ($>$).
For each scenario, we identify unprofitable candidates using the following rules:
\begin{itemize}[noitemsep]
    \item A candidate association tree with a subset of its primitives that is the total set of another candidate (e.g., a candidate performing $SpMM$ and a $GEMM$ is unprofitable compared to another candidate performing only $SpMM$ on the same matrix sizes) 
    \item A candidate with the same matrix primitives as another tree but takes larger matrices as inputs. 
\end{itemize}
\system{} finds unprofitable trees that are common under both scenarios and prunes them away.  
At this point, it cannot prune further without inspecting the input. 
\system{} promotes the remaining association trees to be inspected during the runtime online stage.
It also annotates the candidates when they were profitable ($<,>$) to be used during the final code generation (\Cref{sec:meth_codegen}).

\subsection{Code generation of promoted candidates}\label{sec:meth_codegen}
\begin{figure}[htb]
    \centering
    \includegraphics[width=1\linewidth]{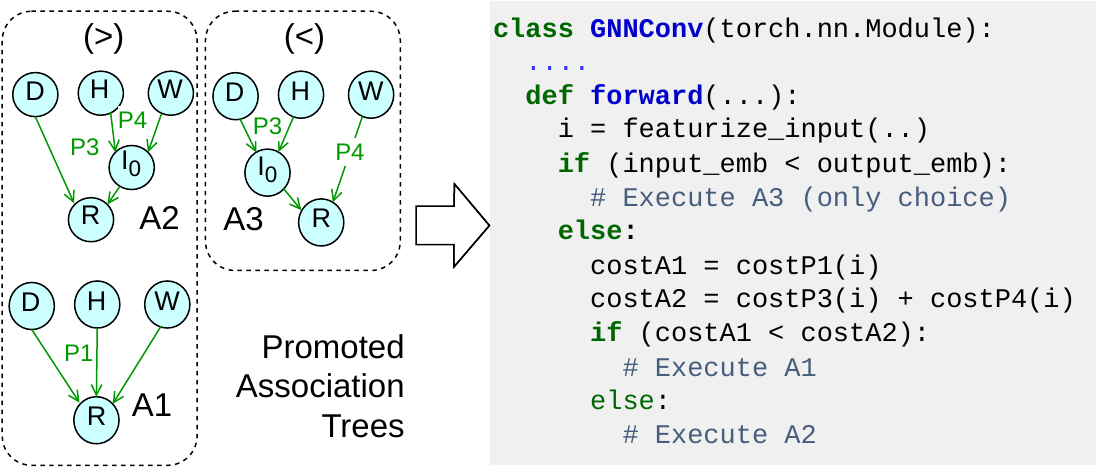}
    \caption{Code-generation of promoted association trees.} 
    \label{fig:codegen_sen}
\end{figure}

The final component of \system{}'s offline stage generates code for the promoted association trees, out of which the best is executed during the online stage. This is achieved by generating conditionally executed code as shown in \Cref{fig:codegen_sen}.
\system{} supports two types of runtime conditions: (1) simpler conditions based purely on embedding sizes and (2) conditions based on more complicated cost models for matrix primitives.

\textit{\textbf{{Conditions using embedding sizes.}}}
\system{} first identifies trees that are profitable at runtime only using embedding sizes.
It does this by categorizing annotated trees that are profitable when either the input embedding size is larger than the output (>), and 2) vice-versa (<).
This avoids the use of the more expensive cost models.
\Cref{fig:codegen_sen} shows an example where the promoted association tree $A3$ is the only one profitable when the input embedding size is smaller.

\textit{\textbf{{Conditions using cost models.}}} 
For the rest of the association trees, \system{} uses cost comparisons using per-primitive cost models that depend on both the embedding size and the input graph. \Cref{sec:prim_cost_models} describes how we develop these cost models. 
We approximate the cost of executing an association tree by the addition of the costs of each primitive.
If multiple association trees will result in the same cost, \system{} selects one tree among them as they are equivalent.

\textit{\textbf{{Code Generation.}}}
Once \system{} has collected the runtime conditions, it progressively generates conditional code to execute the GNN implementation with the best primitive composition. For runtime conditions that require cost comparisons, it embeds the code for cost models for each matrix primitive. 
Once the runtime conditionals are generated, \system{} lowers the matrix primitives of each association tree to kernel calls that are supported by the underlying GNN framework. Figure~\ref{fig:lowering} shows how the final conditionally executed code looks like for an example.

\subsection{Cost models for primitives}\label{sec:prim_cost_models}

\system{}'s cost models predict the cost of a particular matrix primitive, given the input graph and embedding size. We use a graph feature extractor to featurize the input graph characteristics and feed this alongside the embedding sizes to learn simple XGBoost-based~\cite{XGBoost} cost models.

\subsubsection{Graph feature extractor}

We handcraft a set of features from the input graph such as the number of nodes, non-zero values, and density.
We avoided using automatic feature extraction methods such as sparse convolutional networks\cite{Xipeng2018:SparseConvNet} to retain the scalability of \system{} to support large graphs.
NN-based automatic feature extractors do not scale or are too expensive to use in this setting \cite{waco}.

\subsubsection{Lightweight learned cost models}

\system{} uses XGBoost \cite{XGBoost} regression models to predict the input-aware cost of executing matrix primitives. 
\system{} trains these models per each dense and sparse matrix primitive, and target hardware architecture.
This is a one-time cost per target system as the number of matrix primitives used by GNNs are limited in number (g-SpMM for weighted and unweighted graphs, GEMM, row-broadcast, etc.).
Thus, re-training the cost models for each new GNN is unnecessary.
To be input aware, these models use embedding sizes and the features given by the feature extractor for graphs to predict the cost of running (i.e., the time taken) a particular matrix primitive. 
The final cost of GNN execution is the addition of the costs of these primitives. 
\system{} uses a set of profiling data with varying graphs and embedding sizes when training these models (detailed in Section \ref{sec:learning}). 

%% file: paper/implementation.tex
\section{Implementation}
In this section, we explain our training setup and implementation of the components of \system{}. 
These implementations are done in the Python ecosystem.
\subsection{Offline system}
We implemented \system{} to work with both DGL and SeaStar to show its generality. 
\system{} handles the translations from message passing interface to matrix IR (front-end) and association trees to kernel code (back-end) using a simple one-to-one mapping scheme.
We use Python's Abstract Syntax Tree (AST) for this task. 
We implement the rest of the transformations which are core to \system{} using our own classes.
Note that we run the code once when gathering matrix attributes during the translation of the initial code to the matrix IR.
This is an artifact of the dynamically typed Python, where type information is only available at runtime.
We implement the feature extractor and cost models using PyTorch.  
\subsection{Training Lightweight Cost Models}\label{sec:learning}
We collect the training data for the cost models by profiling different matrix primitives on the machines listed below. 
\begin{itemize}[noitemsep, nolistsep,leftmargin=15pt]
    \item CPU - Intel Xeon Gold $6348$, RAM $1$TB, No GPU
    \item GPU - A100 GPU, Intel Xeon Platinum $8358$, RAM $256$GB
\end{itemize}
We source the input graphs to train the model from the SuiteSparse matrix collection. 
Here, we chose a set of undirected graphs ranging from $1$ million to $100$ million non-zero values, which we further varied using sampling.
We gathered approximately $700$ to $8000$ data points for each matrix primitive using this method and by varying the input and output embedding sizes from $32$ to $2048$. 

%% file: paper/primitive_compositions.tex
\section{Case-study: Compositions in popular GNNs}\label{sec:case_prim_comp}

\system{} discovered multiple input-sensitive sparse-dense primitive compositions.
In this section, we dive into the compositions found for two popular GNN models; 
GCN and GAT.
We consider the GNN implementations in DGL\cite{iclr2019:wang:dgl} to be the default.
For clarity, we use the flattened matrix expression in this study.
For both GNNs, \system{} re-discovers the original implementation and identifies an additional primitive composition (\textit{Altn}). 

\subsection{GCN}\label{sec:comp_gcn_main}

The GCN computation has three main calculations: normalization, aggregate, and update. 
The default implementation of GCN  normalizes using two row broadcasts as presented by \Cref{eqn:def_gcn1}.
Here, $\tilde{A}$ is the adjacency matrix of the input graph with self-edges. 
$\tilde{D}$ is the degree of nodes in $\tilde{A}$.
$H^{(l)}$, $W^{(l)}$ are the node embeddings and weights for the $l^{th}$ layer in the GNN.
The result is updated after normalization and aggregation.
Finally, the non-linearity function $\sigma$ is applied.
\begin{equation} \label{eqn:def_gcn1}
\begin{split}
    H^{(l)} =& \sigma((\tilde{A} \cdot (\underbrace{\tilde{D}^{-\frac{1}{2}} \otimes H^{(l-1)}}_\text{row broadcast}) \cdot W^{(l)}) \otimes \tilde{D}^{-\frac{1}{2}})
\end{split}
\end{equation}
Using the initial GCN computation, \system{} generates $12$ association trees.
During the offline phase, \system{} prunes away $8$ using input oblivious rules. The remaining $4$ trees are grouped into $2$ sets (based on the embedding sizes as in \Cref{sec:meth_codegen}), where each set, contains two trees. These compositions are then promoted to the online stage.
We observe two distinct compositions of computing GCN, where only the position of the GEMM is different.

\textbf{\textit{{Dynamic normalization-based primitive composition.}}}
\system{} generates a composition that has a similar normalization mechanism to the default GCN model (as seen in \Cref{eqn:def_gcn1}).
However, it is different due to the update operation (GEMM) being shifted either to the start or the end. 
Notably, this composition enables aggregation without the graph's values.
This makes it possible to use a cheaper \textit{SpMM} (see \Cref{sec:comp_gen}).

\begin{equation} \label{GCN_norm1}
\begin{split}
\tilde{N} &= (\underbrace{\tilde{D}^{-\frac{1}{2}} \cdot \tilde{A} \cdot \tilde{D}^{-\frac{1}{2}}}_\text{SDDMM}) \phantom{XX}
H^{(l)} = \sigma(\tilde{N} \cdot H^{(l-1)} \cdot  W^{(l)})
\end{split}
\end{equation}

{\textbf{\textit{Precomputation-based primitive composition (Altn).}}}
\system{} pre-computes the normalized adjacency matrix $\tilde{N}$ as shown in \Cref{GCN_norm1}.
This computation uses an SDDMM primitive, where the rules mentioned in Section \ref{sec:comp_gen} are used to identify this composition.
This composition is optimal for sparser graphs, as shown by our evaluations in Section \ref{sec:res_dgl}.

\subsection{GAT}\label{sec:comp_gat_main}

\Cref{eqn:def_gat1,eqn:def_gat2} show the default computations of GAT in DGL.
We summarize the underlying computations of attention calculation as a function ($AttenCalc$) to simplify the explanation of the model.
Note that this function uses the updated input embeddings of the nodes ($\Theta = H^{(l-1)} \cdot W^{(l)}$).
${W_A}^{(l)}$ refer to attention weights used during this calculation.
The result is a sparse matrix $\alpha$ containing the necessary attention scores, which is then used for the aggregation and update stages, as seen in \Cref{eqn:def_gat2}.

\begin{equation} \label{eqn:def_gat1}
\alpha^{(l)}_{n \times n} = AttenCalc(\tilde{A},\underbrace{H^{(l - 1)}{\cdot} W^{(l)}}_{\Theta},{{W}^{(l)}_{A}})
\end{equation}

\begin{equation} \label{eqn:def_gat2}
H^{(l)} = \sigma({\alpha}^{(l)}{\cdot}\underbrace{\Theta}_\text{reuse})
\end{equation}

\system{} can only identify $2$ primitive compositions during the offline stage. This happens because computations within \emph{AttenCalc} cannot be re-associated. 
\system{} identifies only one composition that is profitable when the input embedding size is not smaller than the output size (described in \Cref{sec:meth_codegen}) and uses a simple condition to prune during runtime.

\textbf{\textit{{Reuse-based primitive composition.}}}
\system{} re-discovers the default composition of GAT as shown by \Cref{eqn:def_gat2}. 
During aggregation, it reuses the updated embeddings that have already been computed in the attention calculation stage.

\textit{\textbf{{Recomputation-based primitive composition (Altn).}}}
\system{} ignores the reuse of the updated embeddings to generate the primitive composition by using the original embeddings for aggregation. 
However, this requires an additional GEMM computation as shown by \Cref{GAT_reuse_eqn3}. 

\begin{equation}  \label{GAT_reuse_eqn3}
Z^{(l)} = \overbrace{\mathrlap{\phantom{(\alpha^{(l)} \cdot H^{(l - 1)})}}{(\alpha}^{(l)}{\cdot}\underbrace{H^{(l - 1)}) \cdot W^{(l)}}_\text{recomputation}}^{\text{GEMM}}
\end{equation}

%% file: paper/evaluation.tex
\section{Evaluation}

We present our main evaluation of \system{} by comparing it against the default sparse-dense primitive composition for GNN inference and training found in DGL (v.1.1.2). 
For this evaluation, we test across various graphs (1 million - 126 million non-zeros with different sparsity patterns) in both GPU and CPU platforms. 
We used multiple combinations of input and output embedding sizes to showcase different trends in the performance of different input-sensitive compositions. 
In the subsequent sections, we present detailed setups used for evaluation, along with performance results and analyses. 

\vspace{-0.1in}
\subsection{Research questions}
We aim to answer the following questions,

\begin{enumerate}[leftmargin=15pt]
    \item Does \system{} get performance benefits for two popular GNN models (GCN and GAT) on a diverse set of input graphs and embedding sizes? (Section \ref{sec:res_dgl}).
    \item How generalizable is \system{}? 
    Across multiple GNN models and with sampling applied?
    (Section \ref{sec:eval_general}).
    \item How does \system{} compare against other GNN frameworks? 
    Can \system{} see speedups if its techniques are applied there? (Section \ref{sec:other_baselines}).
    \item How beneficial is \system{}'s learned model? (Section \ref{sec:eval_model})
\end{enumerate}
In Section \ref{sec:eval_setup}, we describe our detailed experimental setup.

\vspace{-0.1in}
\subsection{Experimental Setup}\label{sec:eval_setup}

We perform an extensive evaluation to show the benefits of \system{}.
We conduct our main evaluations against the PyTorch \cite{Pytorch} back-end of DGL \cite{iclr2019:wang:dgl}.
Note that this baseline already contains the embedding size aware optimization of the deciding the order of aggregation and update presented in \cite{yan:feat_op_GNN_selection}. 
We evaluate \system{} on the machines listed in Section~\ref{sec:learning}. 
We consider both GPUs and CPUs, as the latter are useful in scenarios where only inference occurs.
We use a single-layer GNN for evaluation since the decisions by \system{} apply to a singular GNN layer. 
As a result, an extension to a multi-layered GNN is achievable by chaining the decisions made for each separate layer. 

We use graphs with multiple variations of non-zero distributions, sourced from domains related to GNN computations.
These graphs are listed in Table \ref{table:eval-data}. 
Among the various types of graphs, we find road graphs (\textit{asia\_osm}), highly dense graphs (\textit{mycielskian17}), and power law graphs (\textit{Reddit}). 
We sourced these graphs from three main locations. 
These were SuiteSparse (SS) \cite{SuiteSparse} -- which provided a plethora of graphs of varying characteristics, DGL and Open Graph Benchmark (OGB)~\cite{hu2020:ogb} -- which provided graphs commonly used in the GNN context. 
The graphs collected for the evaluation were undirected and unweighted, with no overlap with the graphs used for training, 

\begin{table}[ht]
\caption{Graphs used for evaluation}
\label{table:eval-data}
\begin{center}
\begin{small}
\begin{tabular}{c|l|c|c|c}
\toprule
& Graph & Nodes & Edges & Source \\
\midrule
RD & Reddit & 232,965 & 114,615,892 & DGL \\
\midrule
CA & com-Amazon & 334,863 & 2,186,607 & SS \\
MC & mycielskian17 & 98,303 & 100,245,742 & SS \\
BL & belgium\_osm & 1,441,295 & 4,541,235 & SS \\
AU & coAuthorsCiteseer & 227,320 & 1,855,588 & SS \\
AS & asia\_osm & 11,950,757 & 25,423,206 & SS \\
\midrule
OP & ogbn-products & 2,449,029 & 126,167,053 & OGB \\
\bottomrule
\end{tabular}
\end{small}
\end{center}
\end{table}

We use a wide range of embedding sizes for evaluations to showcase the scalability and practicality of \system{}.
In most literature, GNN frameworks are evaluated on a couple of embedding sizes, assuming they would be representative of general workloads.
Usually, these embedding sizes are small powers of 2 (32, 256).
However, we observed a larger range of embedding sizes when inspecting the top ranks of Open Graph Benchmark's leaderboards \cite{ogbnkbrank:shi2023label, ogbnkbrank:zhao2023learning}.
In addition, numerous researches used embedding sizes of 2048 and larger \cite{largeemb:9671372, largeemb:kaufman2021learned, largeemb:recomender}.
This indicates that evaluations found in works introducing GNN frameworks and optimizations are insufficient (see Section \ref{sec:other_baselines}). 
For this reason, we use embedding size combinations ranging from 32 to 2048.
We only evaluate increasing embedding sizes for GAT as this is the scenario in which the primitive composition choice is non-trivial. 

\subsection{Performance Comparison with DGL}\label{sec:res_dgl}

\begin{figure*}[htb]
    \centering
    \subfigure[GCN GPU Inference]{\includegraphics[width=0.49\linewidth]{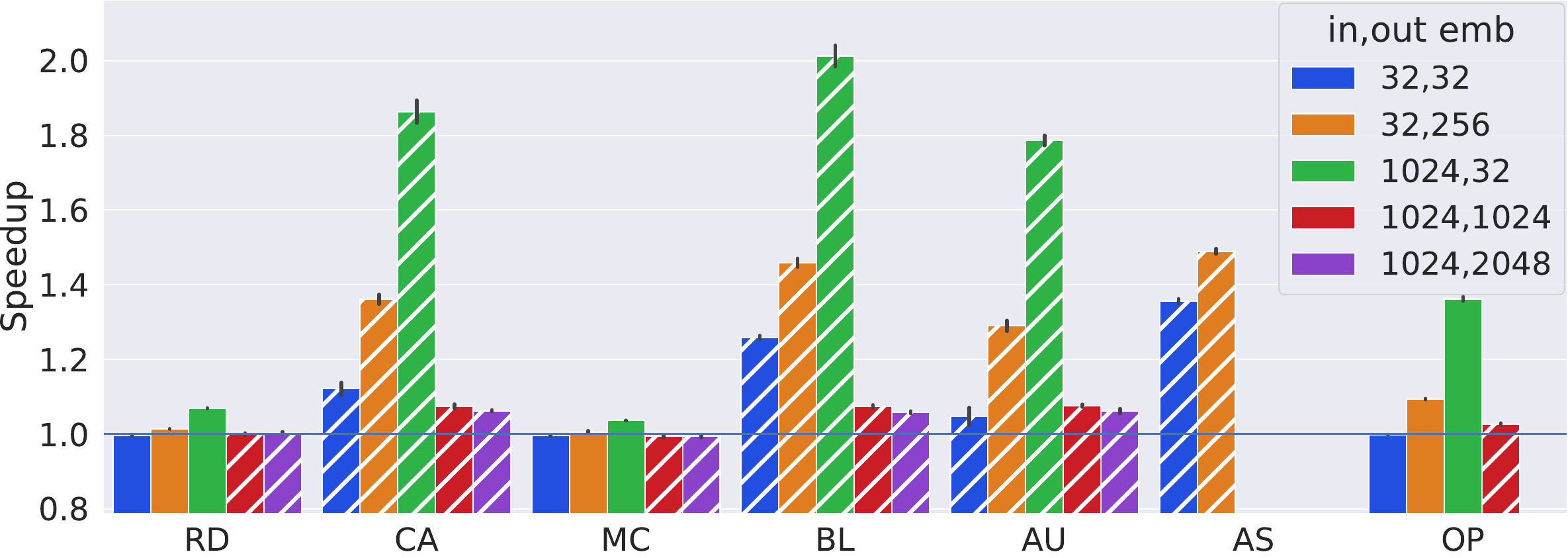}}
    \subfigure[GCN GPU Training]{\includegraphics[width=0.49\linewidth]{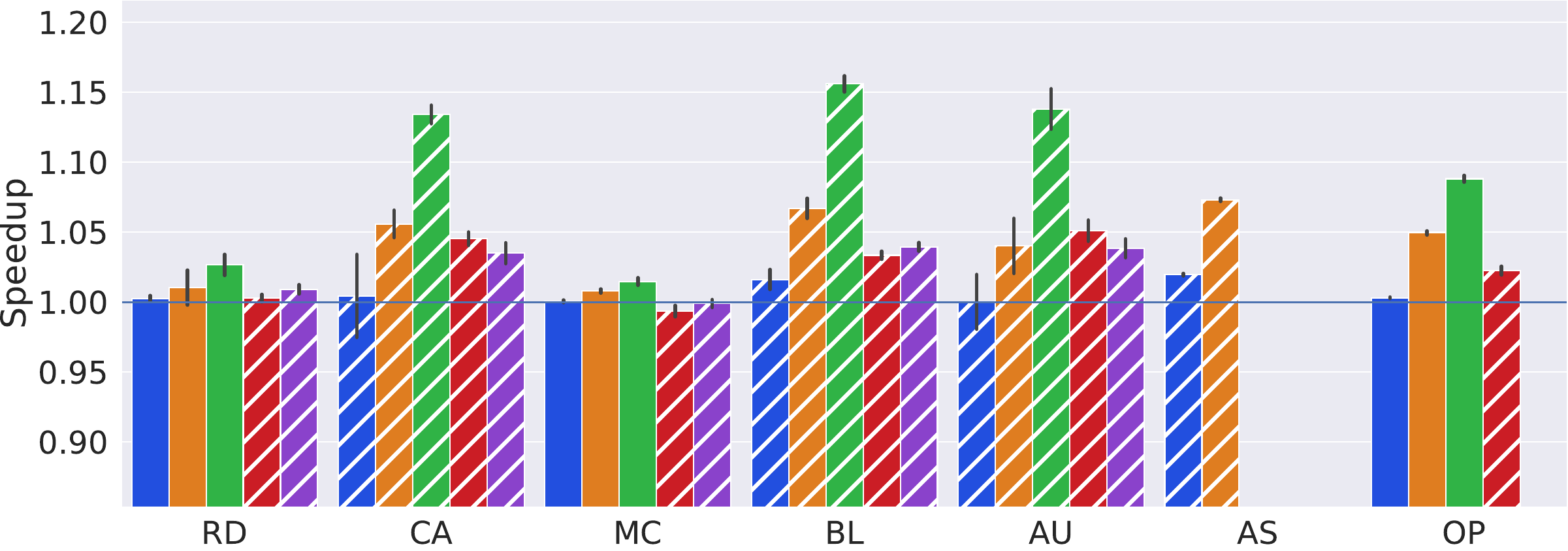}}
    \subfigure[GCN CPU Inference]{\includegraphics[width=0.49\linewidth]{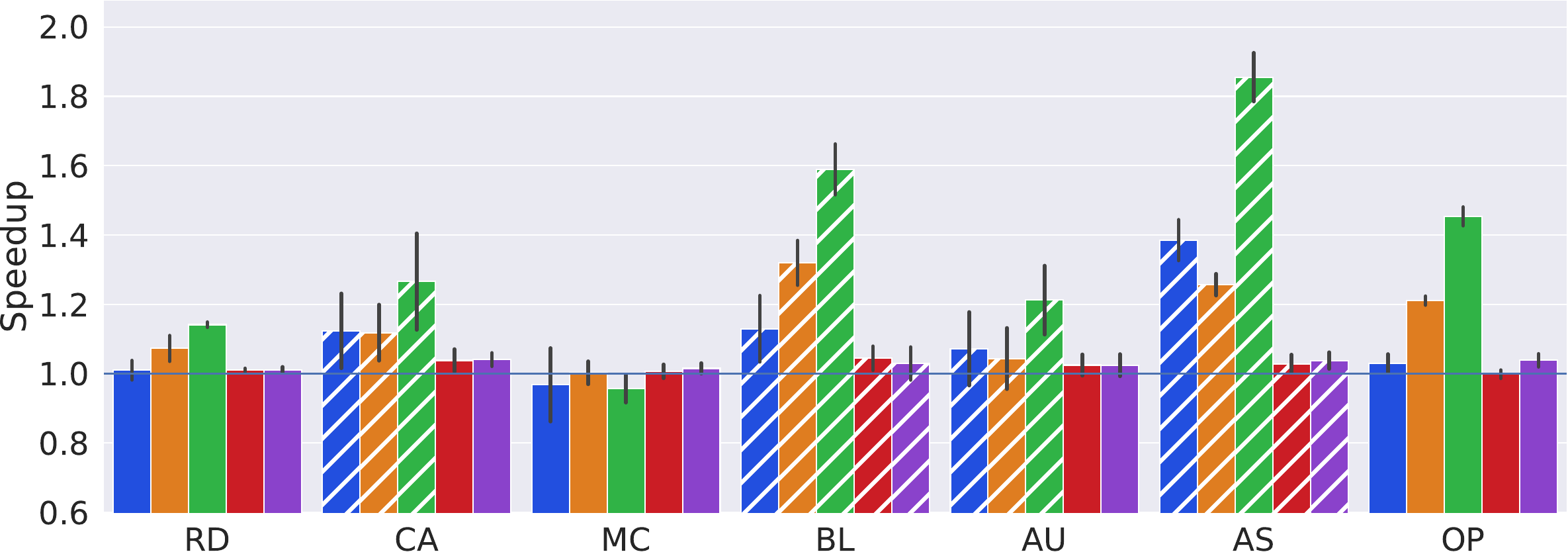}}
    \subfigure[GCN CPU Training]{\includegraphics[width=0.49\linewidth]{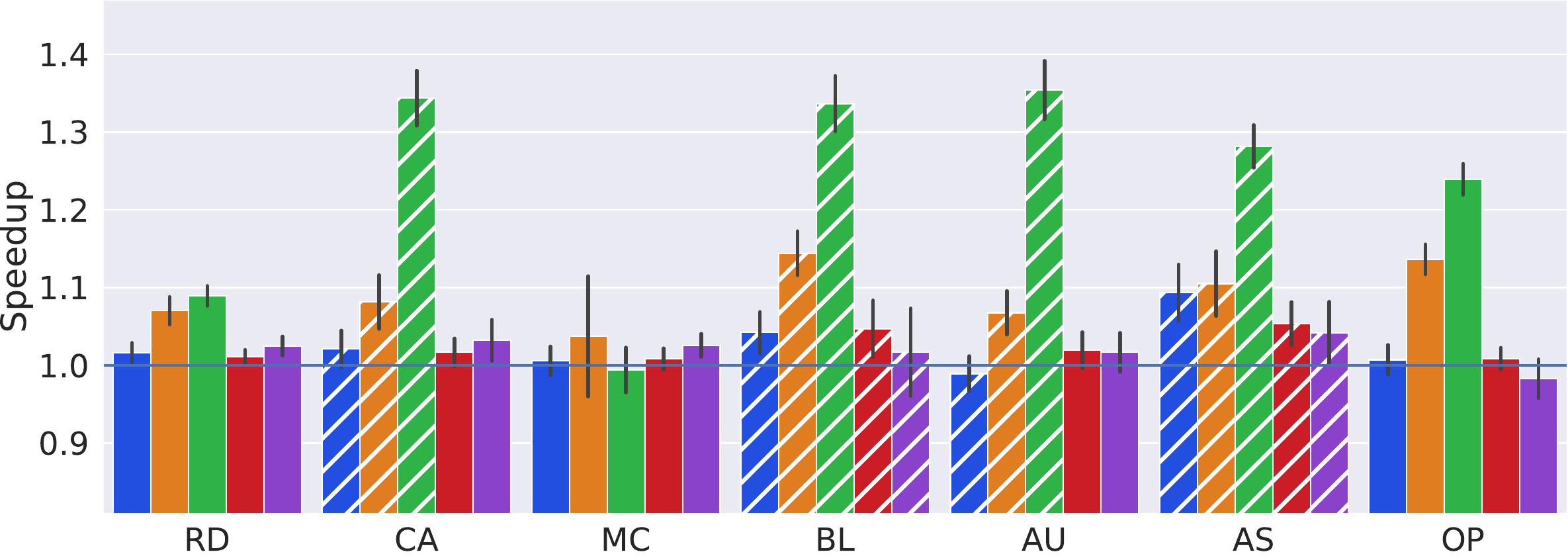}}
    \caption{\system{}'s speedups on DGL - GCN for 100 iterations with runtime overheads. Empty spaces are from OOM failures. Hashed bars denote that \system{} chose \textit{precompute}, and solid bars denote \textit{dynamic} decisions (described in \Cref{sec:comp_gcn_main}).}
    \label{fig:eval_dgl_gcn}
\end{figure*}

\begin{figure}[htb]
    \centering
    \subfigure[GAT GPU Inference]{\includegraphics[width=0.49\linewidth]{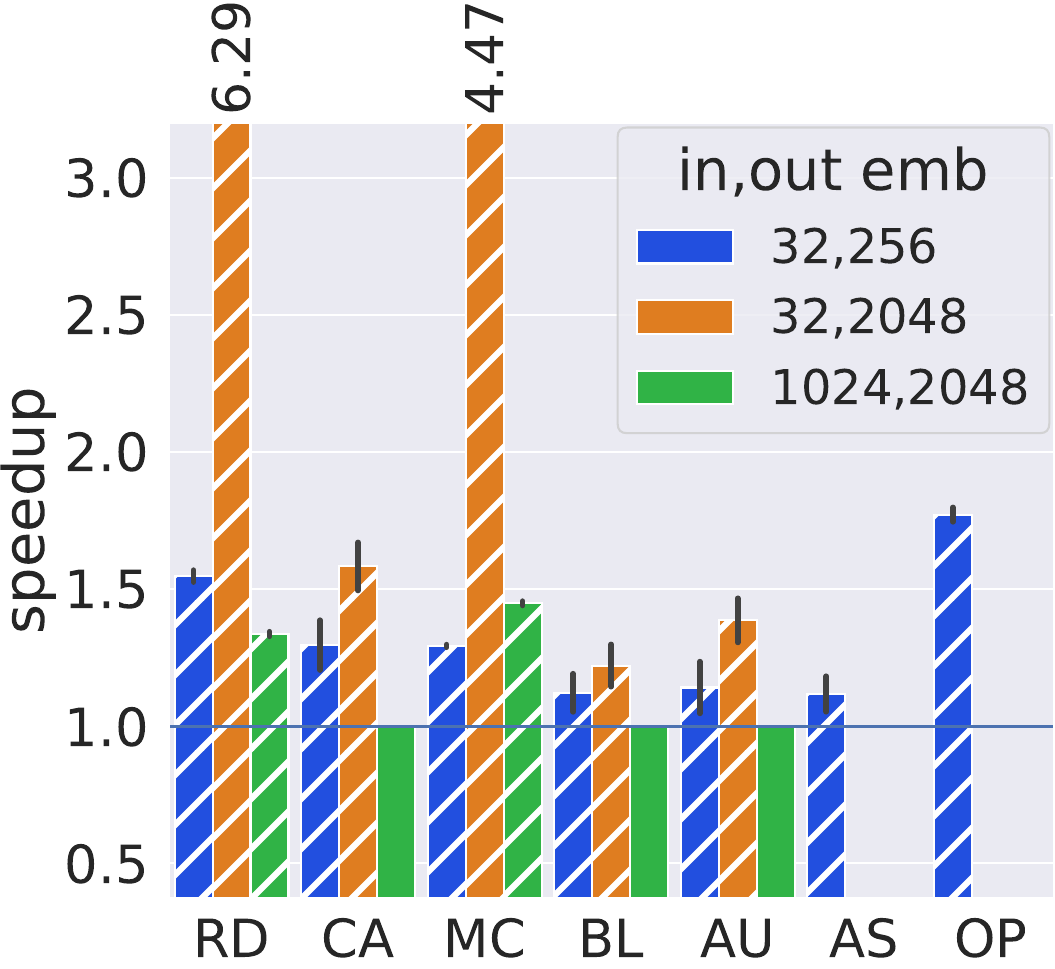}}
    \subfigure[GAT GPU Training]{\includegraphics[width=0.49\linewidth]{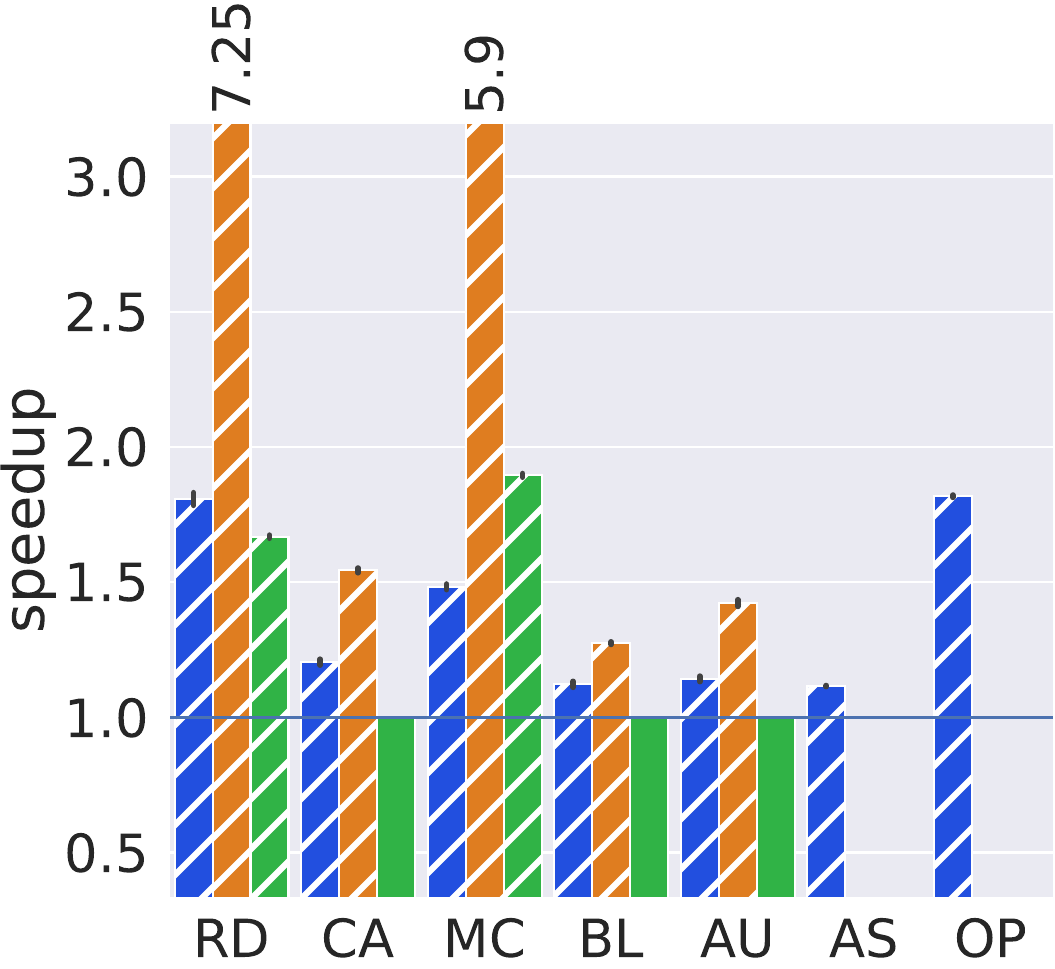}}
    \subfigure[GAT CPU Inference]{\includegraphics[width=0.49\linewidth]{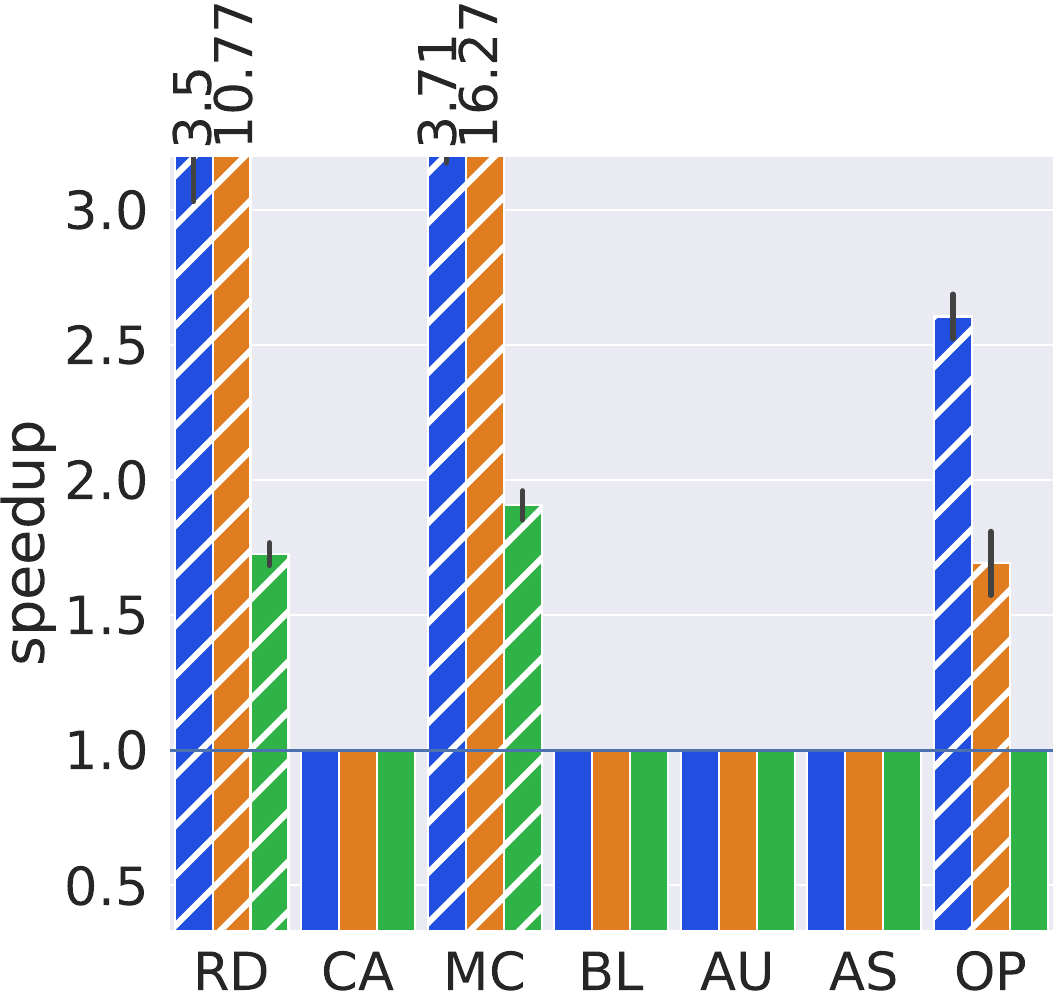}}
    \subfigure[GAT CPU Training]{\includegraphics[width=0.49\linewidth]{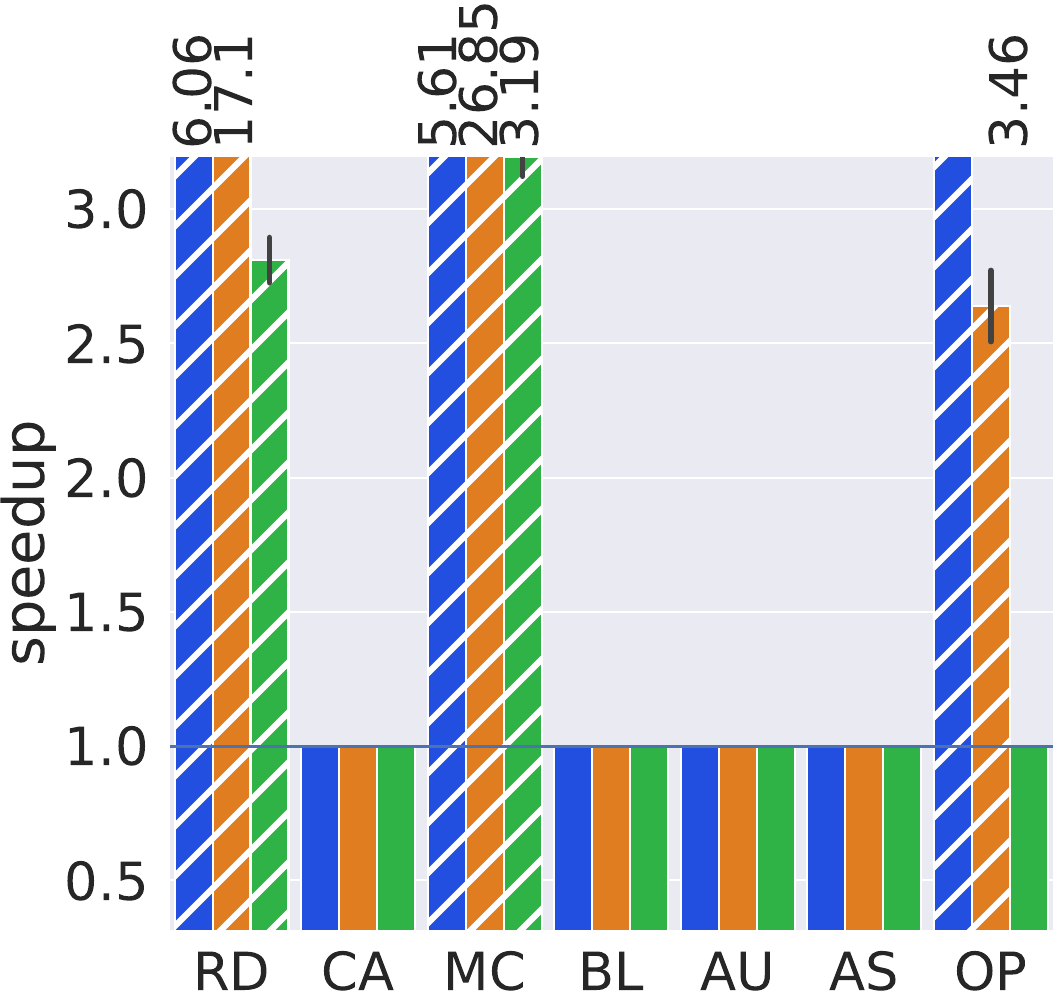}}
    \caption{\system{}'s speedups on 100 DGL-GAT iterations with runtime overheads. Empty spaces are from OOM failures. Hashed bars denote that \system{} chose \textit{recompute}, and solid bars denote \textit{reuse} (described in \Cref{sec:comp_gat_main}).}
    \label{fig:eval_dgl_gat}
\end{figure}

We present the summarized results of our evaluation on both GPUs and CPUs in Table \ref{table:dgl_summary}.
Here, we compare against the default execution in DGL for two popular GNN models, GCN and GAT. 
Focusing on the topic of input-aware execution, we limit the main evaluation to DGL, enabling us to conduct a much more extensive study with various input configurations.
However, we also show \system{} is generalizable to other GNN variants and sampling, as well as compare against other GNN frameworks, in Sections \ref{sec:eval_general} and \ref{sec:other_baselines} respectively.
Notably, through sampling, we are able to support GraphSAGE \cite{gcn:sage} with GCN aggregation.
We run each configuration for 100 iterations and calculate the speedups while considering all overheads. 
We chose 100 to represent the number of times a GNN would typically run, which can range from one iteration to thousands of iterations \cite{waco}.

{\textit{\textbf{Alternative \& Naive execution.}}} 
We also compare the speedups achieved by two different forms of execution for 100 iterations. 
The first form always selects the \textit{Altn} composition. 
The second form is a naive execution strategy(\textit{Naiv}), which runs all selected primitive compositions once, decides the optimal composition among them, and runs that composition for the remaining iterations.
Table \ref{table:dgl_summary} compares the performance of \system{} with the \textit{Altn} and \textit{Naiv} execution models. We observe higher overall speedups with \system{}'s execution model. 

\subsubsection{In-depth comparison}
Figures \ref{fig:eval_dgl_gcn} - \ref{fig:eval_dgl_gat} show an in-depth evaluation of the speedups during inference and training. 
We report trends based on simple features such as the sparsity of the graph and embedding size. 
However, we show in Section \ref{sec:eval_model} that the decisions involved are not trivial.

\begin{table}[th]
\caption{Speedup of \system{} on DGL for 100 iterations}
\vspace{-0.1in}
\label{table:dgl_summary}
\begin{center}
\begin{small}
\setlength{\tabcolsep}{4pt}
\begin{tabular}{l|c|c||c|c||c|c|c}
\toprule
\multirow{2}{*}{\rotatebox[origin=c]{90}{GNN}} & \multirow{2}{*}{\rotatebox[origin=c]{90}{HW}} & Infer(I)/ & SENSEi & SENSEi & \multicolumn{3}{c}{geomean} \\
& & Train(T) & max & min & \system{} & Altn & Naiv \\
\midrule
\multirow{4}{*}{\rotatebox[origin=c]{90}{GCN}} & \multirow{2}{*}{\rotatebox[origin=c]{90}{GPU}} & I & $2.012\times$ & $0.994\times$ &  $1.17\times$ & $1.16\times$ & $\mathbf{1.18\times}$  \\
& & T & $1.156\times$ & $0.993\times$ & $\mathbf{1.04\times}$ & $1.02\times$ & $1.04\times$  \\
\cline{2-8}
& \multirow{2}{*}{\rotatebox[origin=c]{90}{CPU}} & I & $1.85\times$ & $0.945\times$ & $\mathbf{1.11\times}$ & $0.6\times$ & $1.1\times$ \\
& & T & $1.35\times$ & $0.971\times$ & $\mathbf{1.07\times}$ & $0.63\times$ & $1.06\times$ \\
\cline{1-8}
\multirow{4}{*}{\rotatebox[origin=c]{90}{GAT}} & \multirow{2}{*}{\rotatebox[origin=c]{90}{GPU}} & I & $6.294\times$ & $1\times$ & $\mathbf{1.5\times}$ & $1.37\times$ & $1.49\times$ \\
& & T & $7.252\times$ & $1\times$ & $\mathbf{1.61\times}$ & $1.45\times$ & $1.59\times$ \\
\cline{2-8}
& \multirow{2}{*}{\rotatebox[origin=c]{90}{CPU}} & I & $16.274\times$ & $0.998\times$ & $\mathbf{1.64\times}$ & $1.28\times$ & $1.61\times$ \\
& & T & $26.85\times$ & $0.999\times$ & $\mathbf{1.95\times}$ & $1.61\times$ & $1.91\times$ \\
\bottomrule
\end{tabular}
\end{small}
\end{center}
\end{table}

{\textbf{\textit{GCN.}}} \system{} selected \textit{dynamic} for denser graphs when running GCN.
However, for sparser graphs, \system{} selected \textit{precompute}, which led to significant speedups. 
In general, we noticed that sparse graphs have more rows compared to the number of non-zero values. 
An example of a denser graph in this context would be \textit{Reddit}, possessing a density of $2\times10^{-3}$, while a sparser graph would be \textit{asia\_osm}, possessing a density of $1.78\times10^{-7}$. 
Since there are additional row broadcasts in \textit{dynamic}, comparatively more rows are updated in sparser graphs, making it unprofitable.
\system{} manages to accurately identify such points, and makes the correct decision between the primitive compositions. 
In general, we observe higher speedups for \textit{precompute} as the difference between input and output embedding sizes increases.
This is because the larger embedding size makes the GEMM computation have a greater impact on the overall computation.

{\textit{\textbf{GAT.}}} We observe that \system{} selects \textit{recomputation} for the majority of the GAT compositions for GPU.
However, \system{} chose \textit{reuse} for sparser graphs with larger input and output embedding sizes, such as for \textit{belgium\_osm} with embedding sizes $1024,2048$.
This is because the dense computation becomes more prominent as the graph gets sparser, and the embedding sizes get larger.
In such situations, performing the additional GEMM computation to recalculate the updated embeddings becomes too costly.
This is aggravated in CPUs, which are built for more general computations compared to the SIMD architecture of GPUs (where the GEMM computation is more optimized).
In addition, with increasing differences in embedding sizes, the speedups observed from \textit{recomputation} are generally larger. 
This is because the time saved from performing the aggregation with the smaller embedding size becomes more significant.
This is observed for both \textit{Reddit} and \textit{mycielskian17} graphs. 

\textit{\textbf{General Trends.}} 
We observe that speedups for GAT are much greater than GCN. 
This is because the differences in the compositions for GCN are focused on balancing memory use and computation to achieve the best performance, while the compositions of GAT directly lead to significant reductions in computation.
In addition, we see differences in the primitive composition and speedups observed between CPUs and GPUs.
This shows that the optimal primitive composition for a GNN is not solely decided by its input, but also the hardware it operates on.

\textbf{\textit{Overheads.}} 
The slowdowns observed in the evaluations were due to the \system{}'s decision-making process. 
The overheads from graph feature extraction and the composition selection are very minimal -- at most $4.4\times$ of a single GNN iteration for GPU, and $1.1\times$ for CPU. Note that both overheads are incurred only once during runtime.  

\subsection{Evaluating the generalizability of \system{}}\label{sec:eval_general}

\paragraph{Sampling}
In order to check the sensitivity of \system{}'s decision to neighborhood sampling, we evaluated both compositions it discovered for GCN and GAT using $10$ random neighborhood samples of sizes $1000$,$100$, and $10$ on GPU for the \textit{mycielskian17 (MC)} graph. 
We used embedding sizes $(32,256)$, $(1024,2048)$ for GCN, GAT respectively. 
These embedding sizes were selected to visualize changes in the best configuration across our selected sampling sizes. 

\begin{figure}[h]
    \centering
    \subfigure[Sampling for GCN - $(32,32)$]{\includegraphics[width=0.48\linewidth]{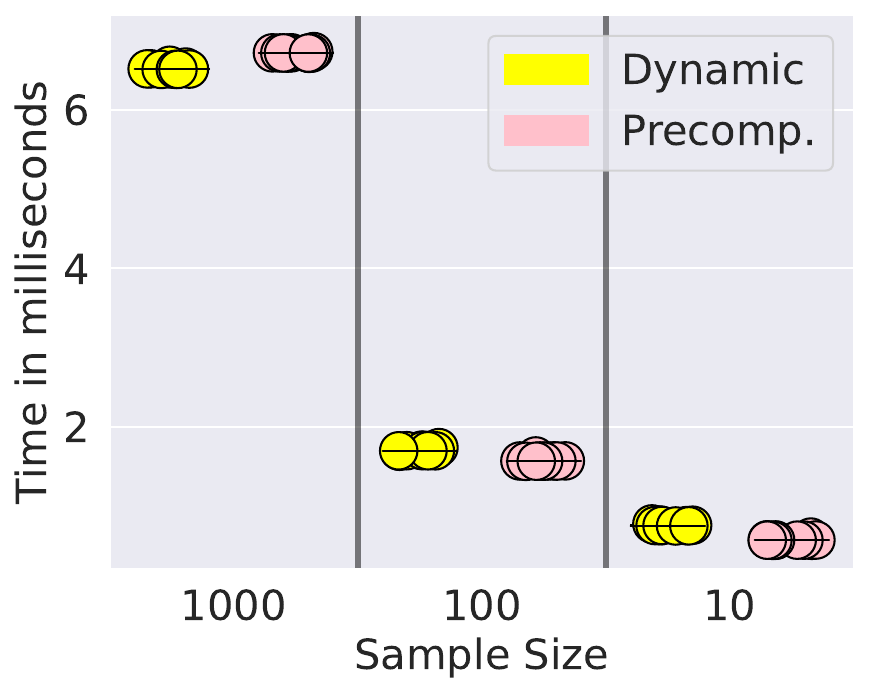}}
    \subfigure[Sampling for GAT - $(1024,2048)$]{\includegraphics[width=0.50\linewidth]{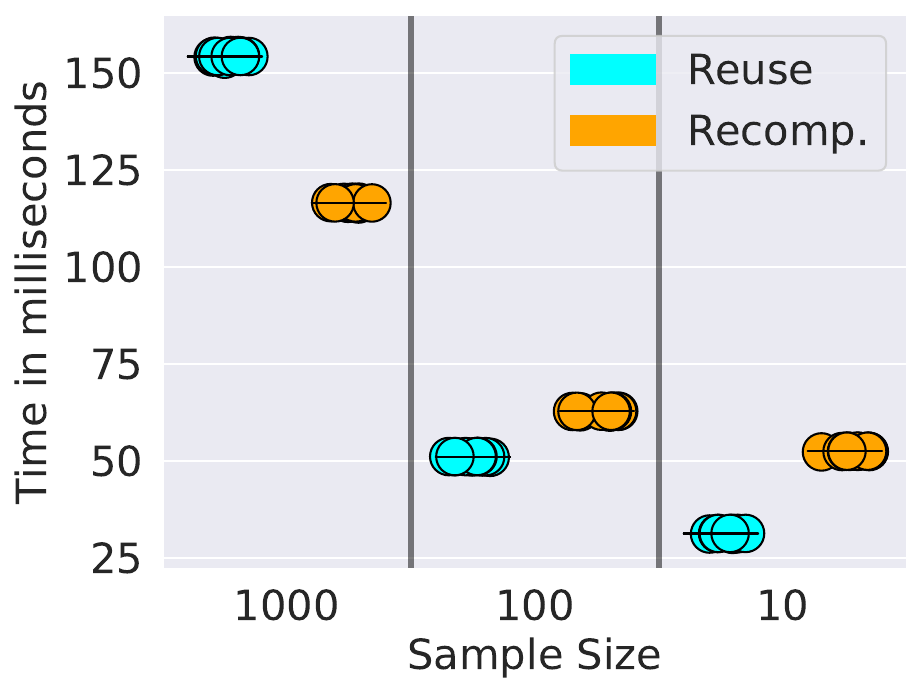}}
    \caption{Scatter plots illustrating the effects of sub-sampling on runtime for \textit{MC} on the GPU. Each composition is run on  $10$ randomly sampled sub-graphs. $(a,b)$ is the embedding sizes. Black lines represent median runtime.} 
    \label{fig:sampling0}
\end{figure}

We make the following empirical observations(as seen in \Cref{fig:sampling0}). First, we see that different random samples of the same node size exhibit minimal variation in runtime. We also noticed that on reducing sample size, \textit{precomputation}(for GCN) or \textit{reuse}(for GAT) becomes more optimal. This aligns with the trends in \Cref{sec:res_dgl}, since a smaller sample size indicates more sparsity.

We evaluate if \system{} is capable of making the correct decision for the sampled graphs with 5 more sampling sizes (ranging from 5000 to 5) in addition to the sampling sizes mentioned previously. 
\system{} only made incorrect decisions when there was little benefit in selecting one composition over the other.
These evaluations show that a single call to SENSEi can be assumed across sampled graphs without the need to inspect the sampled graphs separately and re-run the underlying cost models.

\begin{figure}[h]
    \centering
    \subfigure[Convolution-based variants]{\includegraphics[width=0.51\linewidth]{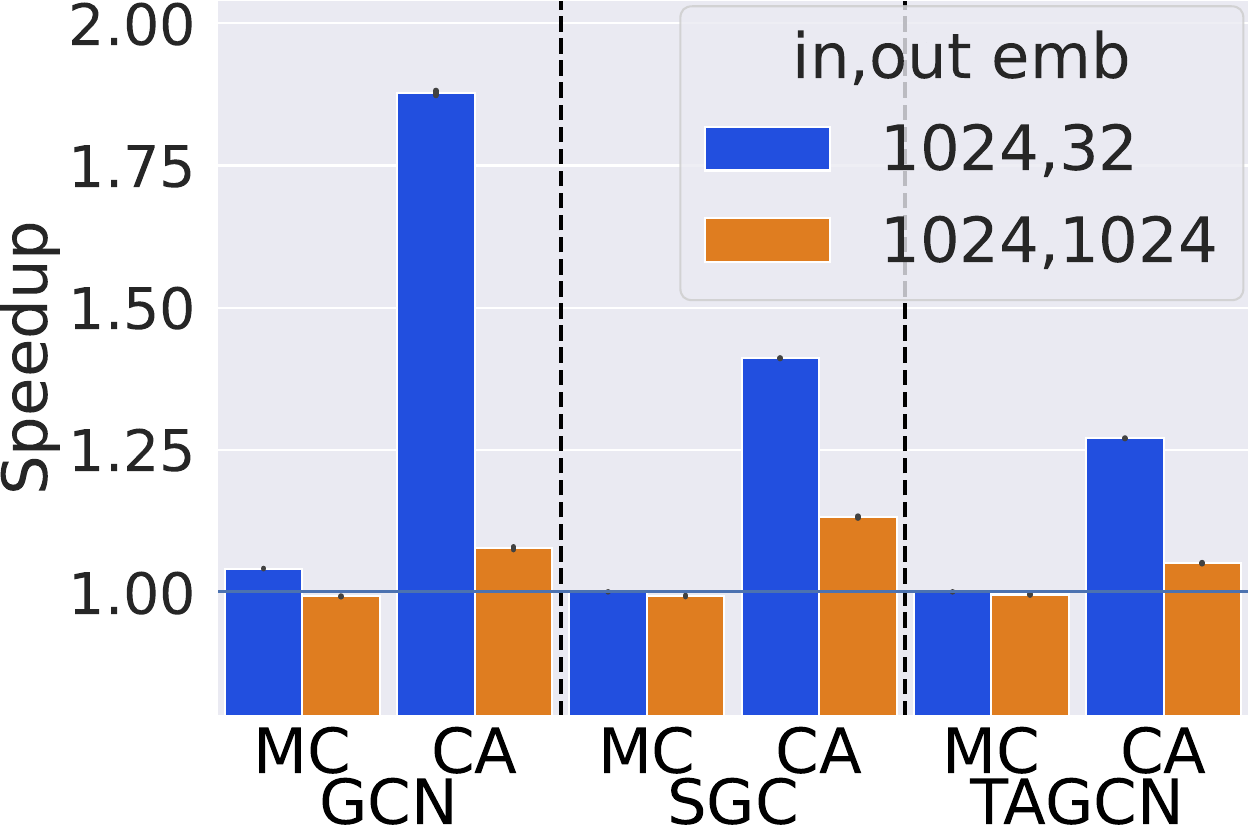}}
    \subfigure[Attention-based variants]{\includegraphics[width=0.48\linewidth]{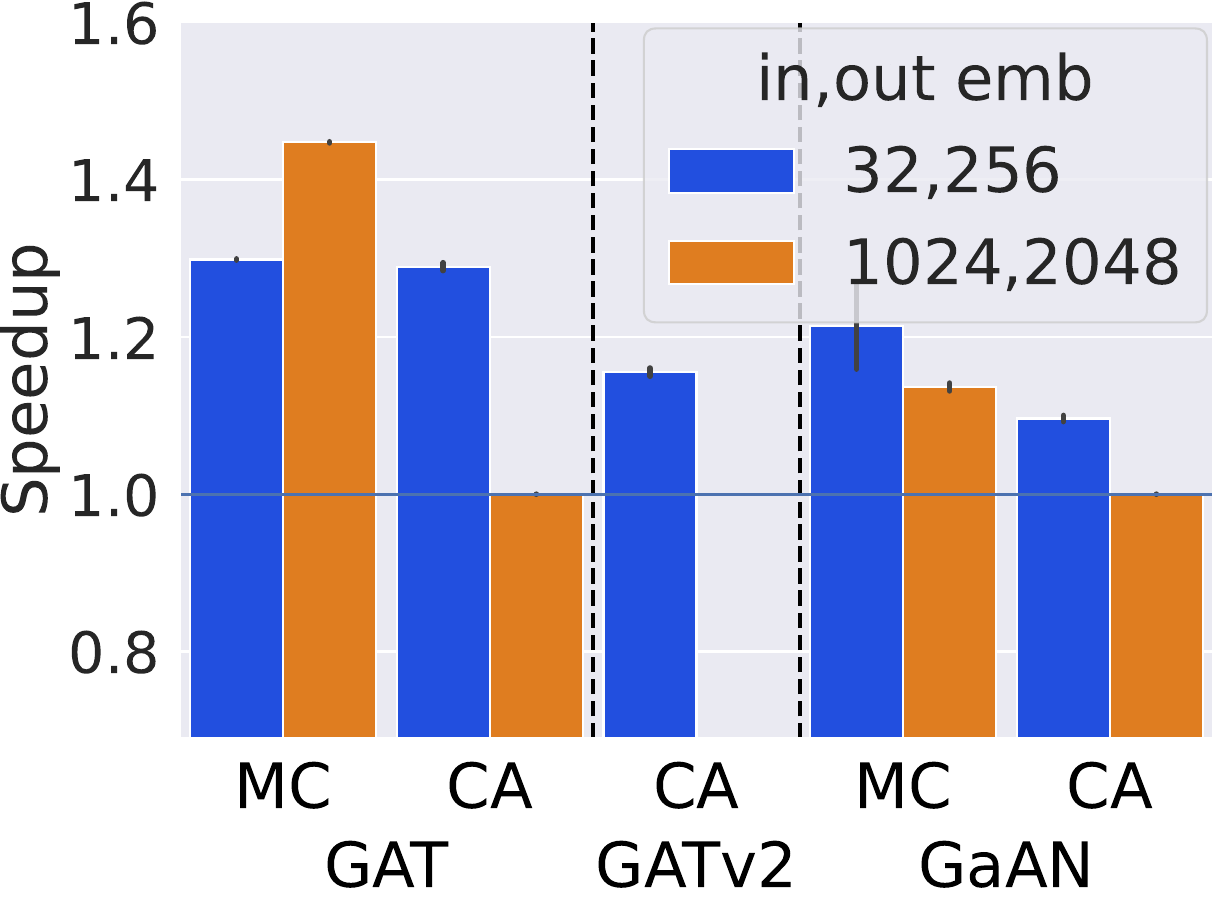}}
    \caption{Effect of \system{}'s decisions on GNNs with similar compositions. Empty spaces are from OOM failures.} 
    \label{fig:feat_variants}
\end{figure}

\textbf{\textit{{Other GNN variants.}}}
We evaluated the decisions made by SENSEi for other GNN variants. 
We observe similar primitive compositions as GCN for Simple Graph Convolution (SGC) \cite{gcn:SGC} and Topology Adaptive Graph Convolutional Networks (TAGCN)\cite{gcn:TAGCN}. 
Meanwhile, we observe similar compositions as GAT in GATv2 \cite{gat:v2} and Gated Graph Attention Network (GaAN) \cite{gat:gaan}. 
As seen in \Cref{fig:feat_variants}, \system{} achieves speedups for multiple GNN models, and is not limited to GCN and GAT. 
To note, the geomean speedups from \system{} were $1.13\times$ and $1.17\times$ for models similar to GCN and models similar to GAT while running only \textit{Altn} gave $1.12\times$ and $1.09\times$, showing \system{} made correct decisions.

\subsection{Comparing speedups with other baseline systems}\label{sec:other_baselines}

\begin{figure}[h]
    \centering
    \subfigure[GCN]{\includegraphics[width=0.55\linewidth]{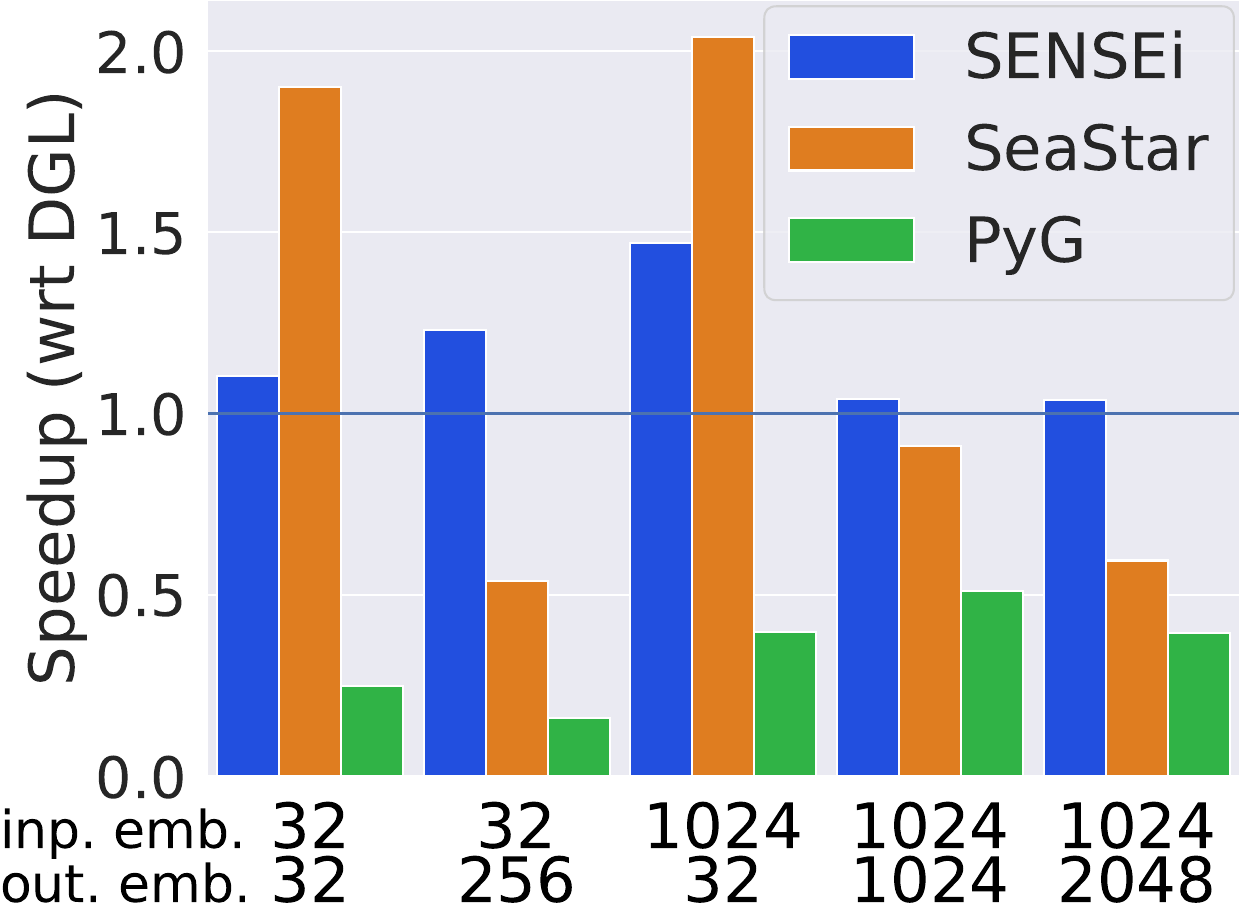}}
    \subfigure[GAT]{\includegraphics[width=0.36\linewidth]{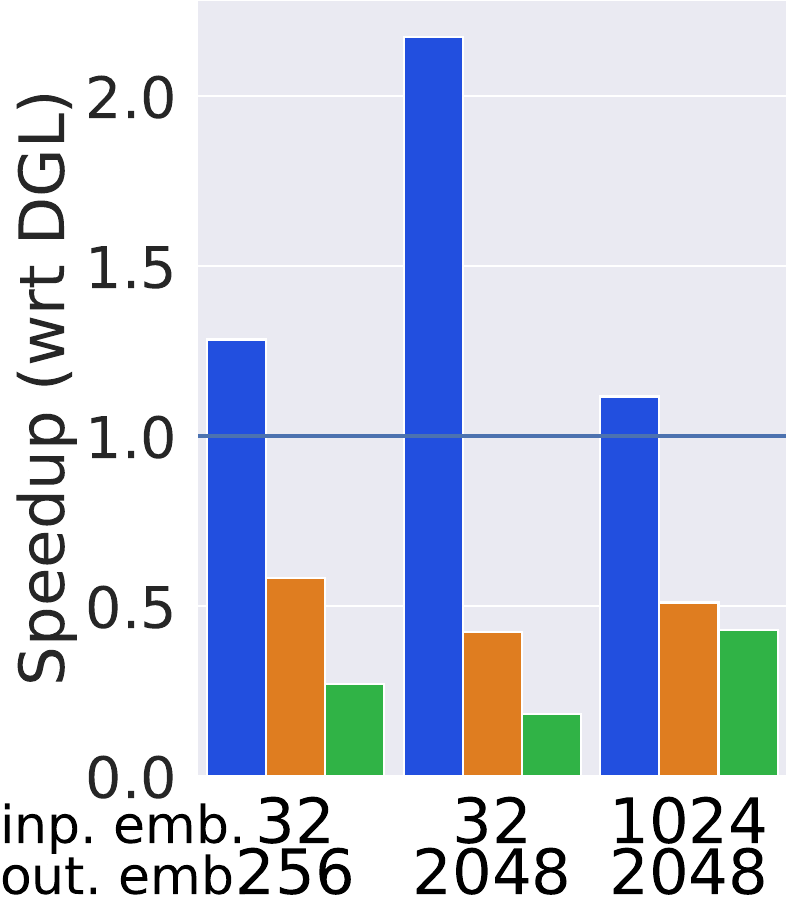}}
    \caption{Geomean of aggregated speedups for \system{}, SeaStar and PyG wrt. DGL per (in, out) embedding size.} 
    \label{fig:other_frameworks}
\end{figure}

\begin{figure}[h]
    \centering
    \includegraphics[width=1\linewidth]{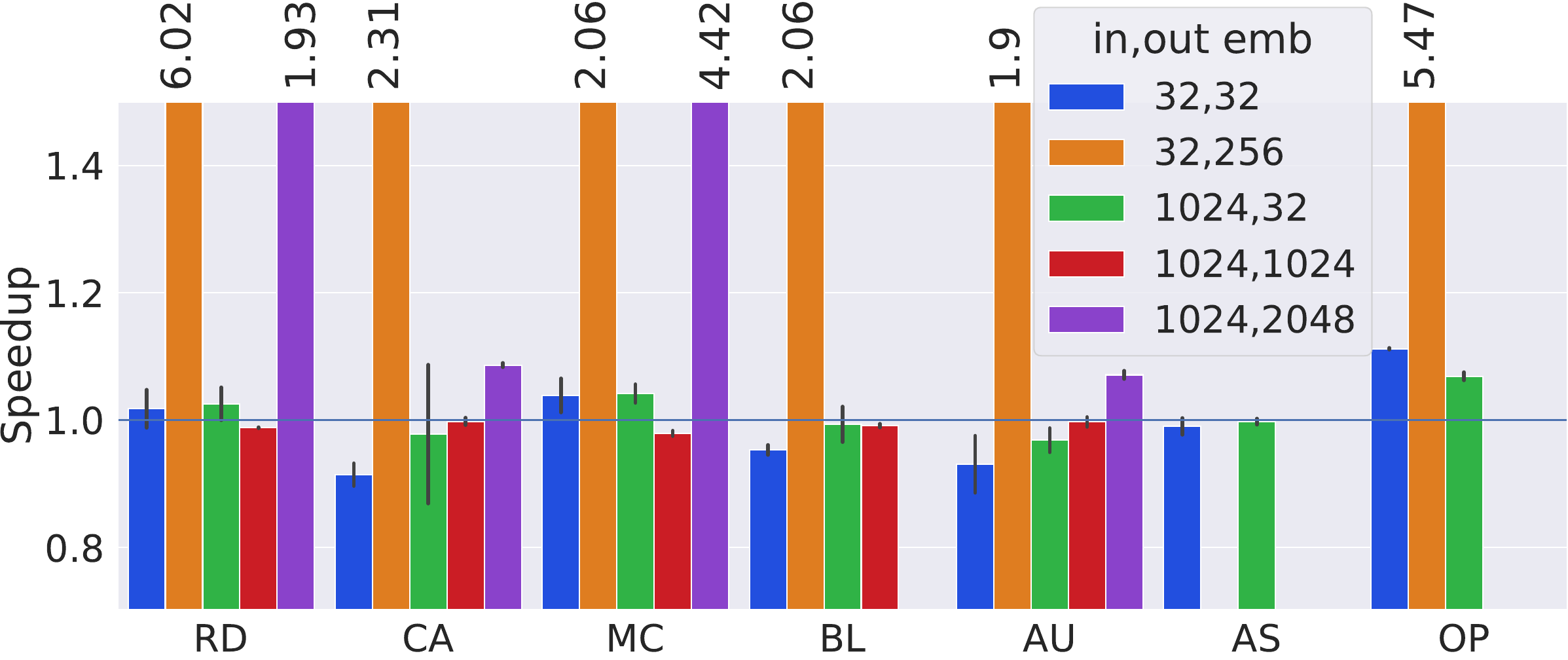}
    \caption{\system{} speedups on 100 SeaStar-GCN inference iterations with runtime overheads.}
    \label{fig:sensei_seastar}
\end{figure}

In this section, we evaluate \system{} against multiple baselines and implement \system{}'s techniques in SeaStar to show that speedups can be observed irrespective of the underlying system.
\Cref{fig:other_frameworks} shows the average speedups of \system{}, PyG \cite{iclr2019:fey:pyg}, and SeaStar \cite{yidi2021:seastar} against the baseline DGL implementations.
Other than the configurations that have an output embedding size of 32 for GCN, Seastar shows performance regressions for the remaining input configurations even against DGL, while PyG shows consistent regressions. 
Our analysis shows that DGL remains a competitive baseline against more recent work such as SeaStar, especially considering scalability.

We implement \system{} in SeaStar to show that the techniques introduced are not limited to a single GNN framework. 
\Cref{fig:sensei_seastar} shows the speedups of \system{} for GCN, where we observe a speedup up to $6.02\times$, with a geo-mean of $1.347\times$ compared to the default SeaStar execution.
The significant speedups achieved for embedding sizes $(32,256)$ and $(1024,2048)$ were due to reordering the GEMM computation based on the input and output embedding sizes.
This further motivates the need for \system{} where such orderings are handled automatically without the need to hand-tune code.
We also see a speedup of up to $22.56\times$, with a geo-mean of $2.64\times$ for GAT with a similar trend to that of DGL.

\subsection{Evaluating the accuracy of \system{}'s cost models}\label{sec:eval_model}

It is non-trivial to select the best primitive composition, as it depends on multiple input factors.
To perform a deeper analysis, we compare the geomean speedup of the optimal configuration (Optimal), with the decisions made by \system{}. 
In addition, we evaluate decisions made by an oracle that selects the best primitive composition independently considering only the underlying hardware architecture (\textit{Arch}), embedding sizes (\textit{Emb}), or the input graph (\textit{Graph}) for both GCN and GAT. 
For example, the \textit{Graph} oracle selects \textit{recompute} as the best for GAT on a given graph if \textit{recompute} is beneficial for a majority of the input configurations (ranging over embedding sizes and target hardware). 

Our evaluations show that \system{} makes more accurate decisions compared to other simpler options.  
The result is presented in Table \ref{table:model_comparison}.
The \textit{Graph} oracle performs the best among the oracle models.
Nonetheless, this shows the need to consider multiple input factors as \system{} does.

\begin{table}[ht]
\caption{Model performance of \system{} against heuristics}
\label{table:model_comparison}
\begin{center}
\begin{small}
\begin{tabular}{l|c|c|c|c|c}
\toprule
GNN & Optimal & \system{} & Arch & Emb & Graph \\
\midrule
GCN & 1.146 & \textbf{1.109} & 1.051 & 0.757 & 1.107 \\
\midrule
GAT & 1.619 & \textbf{1.619} & 1.175 & 1.41 & 1.552  \\
\bottomrule
\end{tabular}
\end{small}
\end{center}
\end{table}

%% file: paper/related_work.tex
\section{Related work}

\paragraph{Input Sensitive Systems}
There has been limited work on systems strictly dedicated to optimizing GNNs based on input. 
GNNAdvisor \cite{ding-gnnadvisor} is one such system where based on the input graph, a set of handcrafted functions are used to identify optimization opportunities. 
Its optimizations are tailored towards GPU-based sparse executions, and the thresholds are heuristically set instead of being learned. 
\cite{qiu:xgboost_format_selection_GNN} is quite similar to \system{} as it proposes a machine-learning model to predict the best data representation for sparse operations in GNNs.
Similarly, WISE \cite{serif:wise} proposes a machine-learning solution to select the best data representation for SpMV.
Works such as ASpT \cite{ASpT} and LAV \cite{serif:LAV} present input-aware sparse optimizations that introduce new sparse representations and are coupled with specialized executions. 
Focusing on graph operations in general, \cite{GPUGraphAutotuner} proposes a solution quite similar to \system{} wherein it presents a learned solution that performs sparse optimizations based on characteristics of the input graph. 
However, this and the aforementioned solutions consider optimizations primarily only on sparse computations. 
By not considering the input-aware interplay of sparse and dense in the context of GNNs, multiple optimization opportunities are missed, as identified by \system{}. 
In contrast, \cite{yan:feat_op_GNN_selection} considers the interplay between sparse and dense computation (which \system{} exploits) but focuses on deciding solely based on the input and output embedding size. Instead, \system{} jointly considers the embedding sizes, the input graph, and the target hardware. 

\paragraph{GNN optimizations}
Graphiler \cite{graphiler}, and SeaStar \cite{yidi2021:seastar} present compiler-based solutions that allow GNN models written in user-defined functions to be converted into highly optimized code based on the computations specified. 
\cite{zhang2021understanding} also considers recomputation as an optimization. 
However, this is in terms of reducing memory utilization of the end-to-end GNN training process and overlooks performance optimizations possible through recomputation as in \system{}. FusedMM \cite{rahman2020fusedmm} and Graphite \cite{josep2022:graphite} also present GNN kernel fusions, albeit specifically on CPU. Notably, the fusion proposed by Graphite is between sparse and dense data components. This relates to \system{} as it considers the interplay between sparse and dense data. As \system{} offers a method of selecting optimizations given the input, all optimizations presented in such systems would compose with \system{}.

%% file: paper/conclusion.tex
\section{Conclusion}
\vspace{-0.1cm}
In this work, we propose a system to exploit the input sensitivity of different primitive sparse-dense compositions in GNN models.
Our system, \system{}, is capable of traversing different primitive compositions and reason about their optimality based on rule and data-driven methods that inspect the input.
Input parameters we consider include the input graph and embedding sizes for a target hardware architecture. 
\system{} allows users to attain speedups compared to the default execution in DGL; a popular GNN framework, over a wide range of graphs and embedding sizes in both GPUs and CPUs. 
\vspace{-0.1in}

%% file: paper/acknowledgements.tex
\section{Acknowledgements}

This work is supported by ACE, one of the seven centers in JUMP 2.0, a Semiconductor Research Corporation (SRC) program sponsored by DARPA.